\documentclass[lettersize,journal]{IEEEtran}
\usepackage{amsmath,amsfonts}
\usepackage{algorithmic}
\usepackage{array}
\ifCLASSOPTIONcompsoc
\usepackage[caption=false,font=normalsize,labelfont=sf,textfont=sf]{subfig}
\else
\usepackage{textcomp}
\usepackage{stfloats}
\usepackage{url}
\usepackage[utf8]{inputenc}
\usepackage{verbatim}
\usepackage{graphicx}
\usepackage{booktabs}
\usepackage{caption}
\graphicspath{{./image/}}
\usepackage{multirow}
\usepackage{subcaption}
\usepackage{color}
\usepackage{tabularx}
\usepackage{indentfirst}
\def\BibTeX{{\rm B\kern-.05em{\sc i\kern-.025em b}\kern-.08em
    T\kern-.1667em\lower.7ex\hbox{E}\kern-.125emX}}
\usepackage{balance}
\begin{document}
\title{Fusion of Infrared and Visible Images based on Spatial-Channel Attentional Mechanism}
\author{Qian Xu, Zheng Chen
\thanks{Manuscript created October, 2020; This work was developed by the IEEE Publication Technology Department. This work is distributed under the \LaTeX \ Project Public License (LPPL) ( http://www.latex-project.org/ ) version 1.3. A copy of the LPPL, version 1.3, is included in the base \LaTeX \ documentation of all distributions of \LaTeX \ released 2003/12/01 or later. The opinions expressed here are entirely that of the author. No warranty is expressed or implied. User assumes all risk.}}

\markboth{Journal of \LaTeX\ Class Files,~Vol.~18, No.~9, September~2020}%
{How to Use the IEEEtran \LaTeX \ Templates}
\maketitle
\begin{abstract}
In the study, we present AMFusionNet, an innovative approach to infrared and visible image fusion (IVIF), harnessing the power of multiple kernel sizes and attention mechanisms. By assimilating thermal details from infrared images with texture features from visible sources, our method produces images enriched with comprehensive information. Distinct from prevailing deep learning methodologies, our model encompasses a fusion mechanism powered by multiple convolutional kernels, facilitating the robust capture of a wide feature spectrum. Notably, we incorporate parallel attention mechanisms to emphasize and retain pivotal target details in the resultant images. Moreover, the integration of the multi-scale structural similarity (MS-SSIM) loss function refines network training, optimizing the model for IVIF task. Experimental results demonstrate that our method outperforms state-of-the-art algorithms in terms of quality and quantity. The performance metrics on publicly available datasets also show significant improvements. 
% Performance metrics on publicly available datasets have also demonstrated significant improvements, such as a 5.28\% improvement in entropy, a \% improvement in mutual information, a 3.48\% improvement in average gradient, and a 5.28\% improvement in standard deviation.
\end{abstract}


\section{Introduction}
\begin{figure*}[htbp]
\centering
\includegraphics[width=1.0\linewidth]{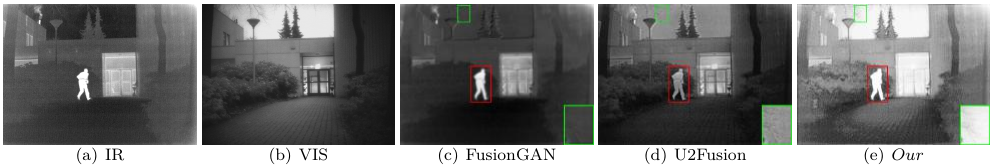}
\caption{Schematic illustration of image fusion. From  left to right: infrared image, visible image, the fusion result of FusionGAN \cite{ma2019fusiongan}, the fusion result of U2Fusion \cite{xu2020u2fusion}, and the fusion result of our proposed AMFusionNet.}
\label{tu1}
\end{figure*}

\IEEEPARstart{I}{mage} fusion is a important problem in the field of computer vision, because  it can provide more informative and reliable information for enhanced understanding and description of a complex scene \cite{song2022medical}, \cite{zang2021ufa}, \cite{liu2022attention}, \cite{ma2019infrared}. One of the most popular image-fusion scenarios is the infrared and visible image fuison (IVIF) \cite{zhang2020vifb}. Infrared and visible images have remarkable complementarities. For example, infrared images can capture thermal radiation emitted from objects, but they lack texture details; In contrast, visible images typically encompass abundant structural information. However, visible images are highly susceptible to environmental conditions, such as heavy fog or low-light situations. Thanks to the complementary characteristics of infrared and visible images, it is worthwhile to fuse them into a composite one.  This integration is beneficial for various downstream visual tasks, including target detection \cite{cao2019pedestrian}, \cite{liu2022target}, 
tracking \cite{zhang2020object}, pedestrian re-identification \cite{zeng2023random}, and semantic segmentation \cite{hou2021generative}.\\
\hspace*{1em}Many methods have been proposed in the past few decades to achieve high-quality fusion from infrared and visible images. These methods can be roughly classified into two categories: traditional methods and deep learning (or neural network)-based method \\
\hspace*{1em}Traditional techniques for image fusion encompass a range of methods, such as multiscale decomposition-based methods (MSD) \cite{gan2015infrared}, \cite{zhu2018novel}, \cite{li2020mdlatlrr}, saliency-based methods \cite{bavirisetti2016two}, \cite{zhang2015fusion}, sparse representation-based methods \cite{liu2016image}, \cite{yang2020infrared}, optimization-based methods \cite{ma2016infrared}, \cite{madheswari2017swarm} and hybrid methods \cite{ma2017infrared}. Among these methods, MSD-based methods have gained significant attention due to their high flexibility and outstanding performance in image fusion. These methods leverage transformation tools, such as the wavelet or pyramid transforms, to decompose the source images into sub-images with different spatial resolutions. They subsequently apply a predefined fusion strategy to fuse the sub-images at corresponding resolution levels. Finally, the inverse transform is applied to obtain the final fused image. While traditional methods have achieved commendable results, the excessive manual intervention has resulted in significant ghosting and noise in the fused images. Furthermore, manually designed feature extractors require substantial time to operate and lack generalization capability.\\
\hspace*{1em}Deep Learning based Methods (DLMs) have achieved remarkable results in various fields, including object recognition \cite{xu2022asymmetric}, \cite{wani2022lowlight}, object segmentation \cite{wang2022freesolo}, \cite{shen2022learning}, and object tracking \cite{yuan2020scale}, \cite{wang2021adaptive}.  Due to the excellent ability for representing feature,  DLMs are becoming the mainstream methods for fusion of visible and infrared images. Based on different fusion architectures, deep learning methods can be roughly divided into three classes: Auto-Encoder-based Methods (AEMs)  \cite{liu2021learning}, \cite{jian2020sedrfuse}, \cite{xu2021drf}, Convolutional Neural Network (CNN) based Methods (CNNMs) \cite{tang2022image}, \cite{long2021rxdnfuse}, \cite{li2021different} and Generative Adversarial Network based Methods (GANMs) \cite{ma2019fusiongan}, \cite{zhou2021semantic}, \cite{xu2019learning}. \\
\hspace*{1em}Although the existing DLMS have achieved satisfactory fusion performance in most cases, there is still room for improvement. Specifically, most of these methods for IVIF use a single size of convolutional kernel, thereby limiting the receptive field of the network and leading to the loss of crucial information during the information extraction process. For instance, AEMs \cite{liu2021learning}, \cite{xu2021classification}, and CNNMs \cite{tang2022image}, \cite{liu2020bilevel}. The IVIF task can be considered an unsupervised task due to the lack of ground truth \cite{hou2021generative}.  Generative adversarial networks (GANs) are well-suited for solving such unsupervised problems by transforming the image fusion process into a game between generators and discriminators. The generator learns the data distribution of the source image, while the discriminator identifies the difference between the generated image and the source image, providing feedback to the network to improve the generation of high-quality images \cite{ma2020ddcgan}. Although GAN-based image fusion networks have demonstrated remarkable advancements and advantages in the field of image fusion \cite{hou2021generative}, \cite{gao2022dcdr}, some textures or salient information may be lost as it is difficult to balance the relationship between the generator and the discriminator. Moreover, most of the GANMs suffers the issue of preserving both the texture information and thermal radiation information from visible and infrared images \cite{yue2023dif}, as show in Figure.\ref{tu1}.(c). Furthermore, inspired by \cite{woo2018cbam}, some researches have successfully incorporated attention mechanisms into the IVIF task, yielding satisfactory outcomes. Nevertheless, all attention network structures implemented in these methods are identical to the modules \cite{ma2020ddcgan}, implying that the channel attention network and the spatial attention network are linked consecutively \cite{wang2023fusiongram}, \cite{zheng2022multi}. Although this architecture has demonstrated notable success in target classification \cite{woo2018cbam}, \cite{meng2022multilayer} and segmentation \cite{wang2020self}, \cite{zhao2021semantic}, its application to the IVIF task might compromise the quality of fused images. This is attributed to potential information loss since each stage's output is directly dependent on its predecessor. As a result, minor inaccuracies or data loss from earlier stages can be exacerbated in subsequent stages. Conversely, the parallel attention mechanism ensures autonomous operation of each stage, safeguarding it from errors of other stages. To date, limited research has investigated the influence of parallel architectures on image fusion performance. To elucidate this shortfall, we reference Fig.\ref{tu1}, which displays the fusion outcomes of U2Fusion \cite{xu2020u2fusion} and FusionGAN \cite{ma2019fusiongan}. Both techniques manifest texture loss and vague boundaries. In contrast, our method circumvents these challenges, yielding images with enhanced contrast and more discernible targets.\\
\hspace*{1em} To address the above-mentioned problems, which often struggle to balance emphasizing prominent targets and preserving texture detail, we proposed AMFusionNet. This innovative network incorporates an attention mechanism, particularly a parallel architecture connecting both spatial attention (SA) and channel attention (CA) networks (PSCNet).  This parallelism facilitates effective information fusion between the attention modules, minimizing information loss and amplifying the network's feature extraction prowess.  Recognizing that human attention is drawn to salient image regions and that texture details enrich our understanding, we emphasize the preservation of such textures.  For enhanced detail retention, we integrate the MS-SSIM loss function into our main loss function, guiding the network's training. While our fusion strategy encompassess three techniques, namely the average-weight method, the $L_1$-norm method, and the mean-filter-operator method.  In summary, our significant contributions are as follows:
% \hspace*{1em}To address the above-mentioned problems, we propose a network based on the attention mechanism. It is worth noting that in this network, the spatial attention network and the channel attention network are connected in parallel. Based on this parallel network structure used for IVIF, effective information fusion can be achieved between the spatial attention module and the channel attention module, reducing information loss during the feature extraction process. Furthermore, this fusion also enhances the network's information extraction capability. Salient regions in images are the focal points of human attention. Texture details within images facilitate our understanding of them; thus, retaining these textures is of paramount importance. To achieve this objective, we incorporated the MS-SSIM loss function \cite{huang2020unet} into our primary loss function to guide the network's training and preserve a greater degree of detail. Regarding the fusion strategies, we used three methods for the fusion layer, including the average-weight method, the $\ell_1-norm$ method, and the mean-filter-operator method. However, these strategies are not used during the training phase.  As a consequence, the fused images generated by our network not only have rich background information that is suitable for human observation, but also retain clear foreground information.\\
\begin{itemize}
  \item We proposed a feature extraction layer that integrates cross-connection of multiple convolutional kernel blocks (MKBlock) in the decoder, enhancing the network's feature extraction capability and expanding its receptive field, while also allowing the network to process features under multi-scale conditions, capturing richer semantic features from images.\\
  \item The network's adaptive capabilities were enhanced by attaching the spatial and channel attention modules after the feature extraction layer, which played a vital role in preserving the texture and thermal radiation information and providing high-quality feature maps for image reconstruction.\\
  \item We introduced a parallel structure of attention mechanism, which can mitigate the information loss during the feature extraction process while enhancing the information fusion between the channel attention branch and the spatial attention path.\\
  \item   Extensive experiments indicate that our methods surpass other representative state-of-the-art (SOAT) fusion methods in both subjective visual description and objective metric evaluation.
\end{itemize}
\section{Related Works}
\subsection{Traditional Algorithms}
Prior to the advancement of deep learning, the predominant approach to image fusion involved manual design of measurement or fusion rules. Traditional fusion methods can be categorized based on different fusion principles such as multiscale transform-based methods \cite{li2011performance}, \cite{lewis2007pixel}, saliency-based methods \cite{zhang2017infrared}, sparse representation-based methods \cite{liu2015general}. \\
\hspace*{1em}Multi-scale transform-based methods are currently one of the most active areas in the field of image fusion. These methods assume that images can be represented by multiple layers at varying resolutions. The multi-scale transformation methods utilized in this area encompass the pyramid and wavelet transforms, such as the Laplacian pyramid \cite{vanmali2017visible} and discrete
wavelet decomposition (DWT) \cite{ren2022fusion}.  These methods involve decomposing the source images into multiple levels, applying specific fusion rules to the corresponding layers, and subsequently reconstructing the target images using the inverse transform of the multi-scale transform. This approach considers the feature variations at different scales, resulting in improved preservation of the image details and texture information. This enables effective preservation of features from both infrared and visible images, thereby enhancing the quality and clarity of the fused image \cite{li2013image}, \cite{ma2016infrared}. \\
\hspace*{1em}Saliency-based methods effectively preserve the vital features of both infrared and visible images in the fused image. These methods enhance object and scene visibility in low-contrast or low-light conditions and improve the detection and recognition performance of computer vision systems \cite{liu2017infrared}, \cite{li2023infrared}. The saliency-based method consists of three main steps: generating saliency maps for both the infrared and visible images, extracting features from both images, and finally combining the features using a fusion rule to create a single fused image that encompasses the most salient features from both images. Extractable features include color, texture, edges, and other visual cues that aid in distinguishing between various regions.\\
\hspace*{1em}Sparse representation, a common method in infrared and visible light image fusion, aims to learn an overcomplete dictionary from high-quality natural images for sparse representation of source images. Fusion rules are applied to sparse coefficients, and the fused image is reconstructed using the learned dictionary based on these coefficients \cite{yang2023latlrr}. This approach is suitable for multi-sensor image fusion and, compared to traditional multi-scale transformations, can derive a more stable and meaningful dictionary from training images \cite{zhang2018sparse}, \cite{ding2018infrared}. \\
\subsection{ Deep Learning Based Methods}
Data-driven deep learning techniques have made significant progress in the field of image fusion by harnessing the robust feature extraction and representation capabilities of deep neural networks. Initially, deep learning-based image fusion methods primarily utilized pre-trained networks as feature extractors for source images \cite{li2018infrared}, \cite{zhang2022infrared}. For instance, Li \cite{li2018infrared} decomposed source images into a base layer and a detail layer, generating weights based on features extracted by deep neural networks. These weights were then applied to the detail layer, while a simple averaging method was employed for the base layer fusion. In  another study by Li \cite{li2020nestfuse}, a fusion approach utilizing nested connections was introduced. This approach effectively leveraged multi-scale information by merging features across different levels, thereby preserving both spatial and texture details. 
\hspace*{1em}Considering the suitability of Generative Adversarial Networks (GANs) for image fusion tasks, Ma \emph{et al} \cite{ma2019fusiongan} presented a fusion approach for infrared and visible light images utilizing GANs. By harnessing the generative power of GANs and their capability for automatic feature learning, this approach facilitated the adaptive extraction of vital information during the fusion process. However, GAN-based methodologies have encountered significant issues that require further resolution. These methods fuse source images as the input image and rely solely on a discriminator for adversarial training, resulting in a deficiency of local detail and indistinct target edges in the fusion image. In addition to GAN-based approaches, several general image fusion frameworks have been proposed. For instance, Xu et al \cite{xu2020u2fusion} proposed a framework designed for multiple fusion tasks, including multi-exposure, multi-focus, and multi-modal situations. This framework evaluates the significance of the source images through feature extraction and information measurement, determining an adaptive information preservation degree. By training the network to uphold an adaptable similarity between the fused outcomes and the original images using this adaptive degree, the framework achieves effective fusion. However, this approach exhibits a weak capacity for preserving the thermal radiation information of infrared images, resulting in slightly blurred edges in the fused image. To address these limitations, Ma \cite{ma2020ddcgan} introduced an end-to-end universal fusion framework consisting of two discriminators and one generator. By designing specific loss functions and establishing a game between the generator and discriminators, this framework successfully preserves both the thermal radiation information from the infrared images and the texture details from the visible light images.\\
\hspace*{1em}Currently, image fusion using Auto-Encoder (AE) has become a mainstream approach. AE which are unsupervised neural networks, are capable of learning efficient data representations by encoding and decoding input data. One notable and innovative AE-based IVIF method is DenseFuse \cite{li2018densefuse}, which utilizes MS-COCO  \cite{lin2014microsoft} for pretraining the autoencoder and employs distinct fusion strategies (addition and $L_1$-norm) to perform feature fusion. Additionally, Xu et al. \cite{xu2022cufd} introduced a dual-branch encoder-decoder network architecture. In the feature extraction stage, this network efficiently extracts both shallow and deep features, allowing for simultaneous extraction of infrared radiation and texture information. The source images undergo decomposition into shared and distinct information by considering the similarities and differences between the infrared and visible images. Various fusions are then performed during the feature reconstruction stage based on the significance of the feature maps. To further optimize the performance of IVIF  networks based on AE, some researchers have incorporated attention mechanisms. These mechanisms typically encompass both channel attention and spatial attention branches. In this manner, the network can focus more on salient target areas of source images, thereby enhancing its feature representational capability.   For example, FusionGRAM \cite{wang2023fusiongram} combines dense connection networks \cite{huang2017densely} with attention mechanisms to enhance the network's feature extraction capability and adaptively allocate more weight to salient regions. Additionally,  Xu \emph{et al} \cite{xu2022multi} proposes a network that utilizes a combination of skip connection networks and  attention mechanisms. The attention module comprises three attention networks that employ different convolutional kernels to facilitate feature fusion across multiple scales. This enables the network to preserve detailed information more effectively. Despite the positive outcomes of attention-based IVIF networks, the sequential structure of their spatial attention network and channel attention network might lead to the loss of vital information during image fusion. \\
% \hspace*{1em}Although existing methods yield satisfactory results, they often struggle with achieving a balance between emphasizing prominent targets and preserving texture information. To tackle this issue, we propose AMFusionNet, which incorporates several technical advancements compared to conventional approaches. First, AMFusionNet utilizes multi-convolutional kernel cross-connections in the feature extraction layer to enhance the network's receptive field and extract valuable information more effectively. Second, employing a parallel architecture for both the spatial attention network and the channel attention network. Lastly, by incorporating an MS-SSIM loss function into the overall loss function, the fused images achieve a balance between preserving visible light and infrared image information. This results in fusion images that are better suited for human observation.
\section{Network and loss function}
In this section, we shall present the details of the fusion strategy and the overall framework of the proposed network.
\subsection{Network Architecture}
\begin{figure*}[htbp]
\centering
\includegraphics[width=1.0\linewidth]{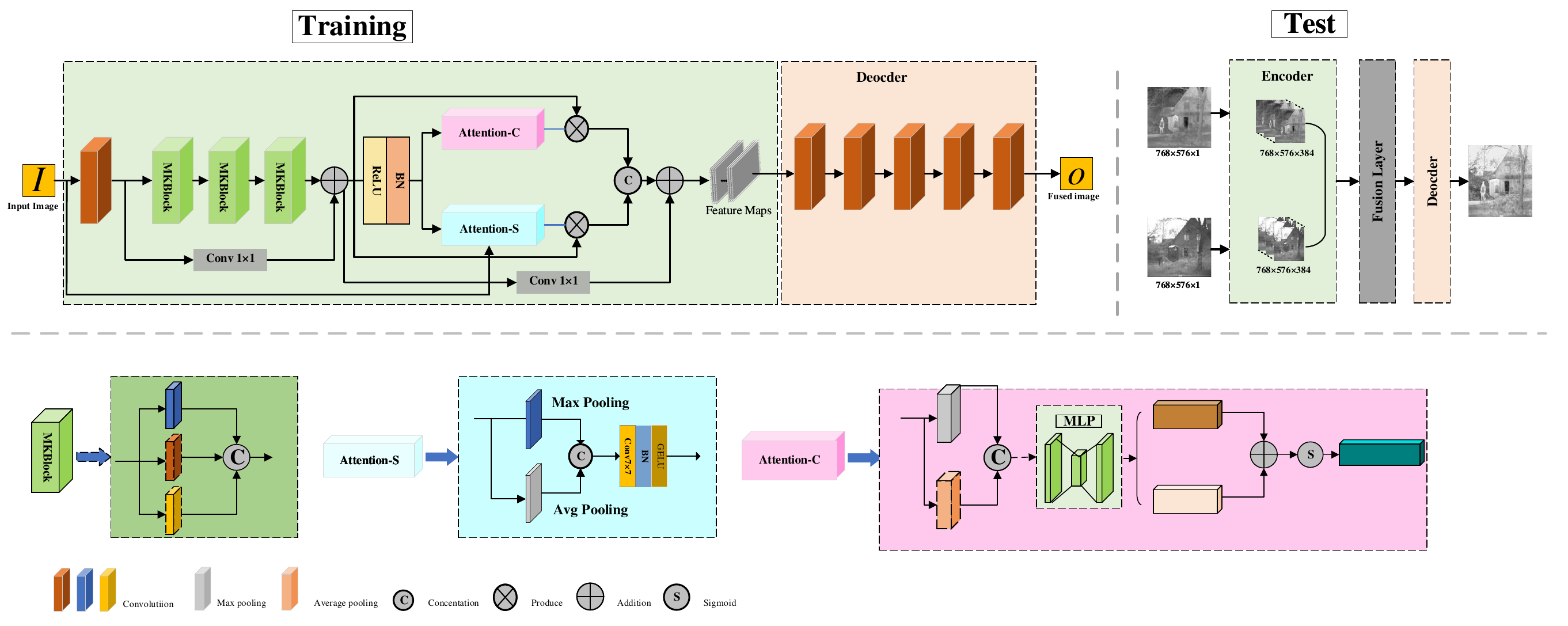}
\caption{The architecture of the proposed infrared and visible image fusion network based on salient target detection}
\label{jiegoutu}
\end{figure*}

As shown in Fig.\ref{jiegoutu}, the AMFusionNet consists of two parts: one is an encoder and the other is a decoder.\\
\hspace*{1em} The encoder employs a CNN layer for preliminary feature extraction, followed by three MKBlock layers and PSCNet modules. The initial CNN layer, which uses a $3 \times 3$ kernel size and is complemented by a Rectified Linear Unit (ReLU) activation layer, extracts foundational features from the source images.  The MKBlock features multiple interconnected convolutional kernels of sizes $3\times3ji$, $5\times5$, and $7\times7$. As such, each input map within the MKBlock is processed by various convolutional kernels, yielding feature maps representing edges, textures, and shapes.  These feature maps can extract different features, such as edges, textures, and shapes. These diverse features are then integrated using cross-connections, facilitating a comprehensive and varied feature representation. The PSCNet is a parallel attention network, where the spatial attention emphasizes spatial features such as edges and textures, while the channel attention emphasizes channel-level information dynamically adjusted based on complementary data across channels. This parallel structure of the attention mechanism can fully integrate the output features of the channel attention network and the spatial attention network, thereby reducing information loss during the feature extraction process. To counteract potential feature loss from numerous network layers, a residual structure is incorporated, which combines the final MKBlock's output with the attention mechanism layer's outcome after a convolutional process.\\
\hspace*{1em} The decoder serves as a feature reconstructor, comprising five convolutional layers with $3\times3$ kernel sizes. The initial four layers incorporate Batch Normalization (BN) and Parametrized ReLU (PReLU) as their activation functions. In contrast, the fifth layer employs the hyperbolic tangent (Tan) as its activation function. \\
% \hspace*{1em}Traditional convolutional neural network ($\bf CNN$) methods usually use fixed-size kernels for convolution, and they generally lose parts of the target's feature information, resulting in unsatisying fused image. To prevent the loss of feature map information due to convolution operations, $ \bf MK$ design a convolutional methodology with a multichannel parallel topology, it uses convolutional kernels of different sizes (the kernel size is $3\times3$, $5\times5$, $7\times7$ respectively) to extract the multikernel-size deep features of the infrared and visible target. Additionally, the $\bf MK$ module can realize information interaction through cross-connect operation among the three convolution streams, and generates the feature maps $\boldsymbol{F_{mk}}$ of source image. Then, the network module concatenates the three kind of feature maps along channels. The concatenated feature maps pass through $\bf SCAM$ module generates feature maps which is consisted of two kind of attention module, one is channel attetion module, and the other one is spatial attention module. To prevent the detail information of the feature maps from being lost after the multiple convolutionals operator, we fed the feature maps to convolution layer (kernel size is $1\times1$) generates feature maps $\boldsymbol{F_{scam}}$, and add the feature maps to the $\bf SCAM$ module outputs get feature maps $\boldsymbol{F_{fused}}$. At last, feature maps $\boldsymbol{F_{fused}}$ pass through decoder to recover the original image.\\ 
\begin{table}
	\begin{center}
		\caption{Network configuration.\textbf{Conv3, Conv5, Conv7 denotes the convolutional layer with $3\times3$, $5\times5$, $7\times7$, respectively. }}
		\label{tab1}
		\centering
		\resizebox{\linewidth}{!}{ 
			\begin{tabular}{| c | c | c | c | c | c | c | c |}
				\hline
				\multirow{2}{*}{Module} & \multirow{2}{*}{Layer} & \multirow{2}{*}{Size} & \multirow{2}{*}{Stride} & Input & Output & \multirow{2}{*}{Padding} & \multirow{2}{*}{Activation}\\
				&&&& Channel & Channel &&\\
    				\hline 
                \multirow{2}{*}{PSCNet}&SA&-&-&384&384&-&- \\ \cline{2-8} 
				&CA&-&-&384&384&-&- \\ \hline
				\multirow{2}{*}{Encoder}&MKBlock &-&-&-&-&1&- \\ \cline{2-8}
				&PSCNet &-&-&-&-&1&- \\ \hline
				\multirow{5}{*}{Decoder}&Conv3 &3&1&768&384&1&PeLU \\ \cline{2-8}
				&Conv3  &3&1&384&192&1&PeLU \\ \cline{2-8}
				&Conv3  &3&1&192&96&1&PeLU \\ \cline{2-8}
				&Conv3  &3&1&96&48&1&PeLU \\ \cline{2-8}
				&Conv3  &3&1&48&1&1&Tan \\ \hline
				\multirow{3}{*}{MKBlock}&Conv3&3&1&16,96,192&32,64,128&1&PRLU \\ \cline{2-8}
				&Conv5&5&1&16,96,192&32,64,128&1&PRLU \\ \cline{2-8}
				&Conv7&5&1&16,96,192&32,64,128&1&PRLU \\ \hline
				\multirow{2}{*}{SA}&Conv3&3&1&16,96,192&32,64,128&1&PRLU \\ \cline{2-8}
				&Maxpooling&-&1&16,96,192&32,64,128&1&PRLU \\ \cline{2-8}
				&Averagepooling&-&1&16,96,192&32,64,128&1&PRLU \\ \hline
    		\multirow{2}{*}{CA}&Conv3&3&1&384&384&1&PRLU \\ \cline{2-8}
				&Maxpooling&-&-&384&1&1&PRLU \\ \cline{2-8}
				&Averagepooling&-&-&384&1&1&PRLU \\ \hline
			\end{tabular}
		}
	\end{center}
\end{table}
\subsection{Loss Function}
The construction of the loss function is crucial for the proficient reconstruction of input images. Numerous loss functions have been proposed in existing literature, merging pixel-based loss with the Structural Similarity Index Measure (SSIM) loss \cite{long2021rxdnfuse}, \cite{wang2004image}. These functions are mathematically articulated as:
\begin{equation}
	\label{equ1}
	L_{pixel}=\frac{1}{BCHW}\Vert O-I \Vert_2
\end{equation}
\begin{equation}
	\label{equ2}
	L_{ssim}=1-SSIM(O-I)
\end{equation}
\begin{equation}
	\label{equ3}
	SSIM(O,I)=\frac{(2\mu_O\mu_I+C_1)(\sigma_{OI}+C_2)}{(\mu^2_{O}+\mu^2_{I}+C_1)(\sigma^2_{O}+\sigma^2_{I}+C_2)}
\end{equation}
where $O$ and $I$ are output and input images, respectively, $\Vert \bullet \Vert_2$ is $L_2$ Euclidean norm, $B$ respresents the batch size; $C$ is the number of channels of image $O$, $I$; $H$ and $W$ are height and width of image. $\mu_X$ and $\mu_Y$  represent the mean pixel values of image $0$ and $I$, respectively, respectively.  $\sigma_X, \sigma_Y$ denote the standard deviations of the two images, respectively.  $\sigma_{XY}$ represent the covariance of the two images, which is a measure of the extent to which the corresponding pixel values of the images change together. $C_1$ and $C_2$ are small positive constants.\\
\hspace*{1em}Since the $L_{pixel}$ loss function solely minimizes the Euclidean distance between pixel values in input image $I$ and output image $O$, it does not account for the semantic relationships among pixels. Moreover, it is prone to challenges related to the loss of fine details and an excessive degree of smoothnes. To address these limitations, we propose incorporating multi-scale SSIM (MS-SSIM), and $L_1$-norm loss functions. The two loss functions enable the fused image to retain color and luminance  from the input images while enhancing the contrast of the fused image. The MS-SSIM loss function $L_{ms-ssim}$, and $L_1$-norm loss function are computed as follows:
\begin{equation}
	\label{equ5}
	L_{MSSSIM}=1-l_M(O,I)\cdot\prod_{j=1}^Mcs_j(O,I)
\end{equation}
\begin{equation}
	\label{equ6}
	L_{l_1}=\frac{1}{HW}\left \|O-I \right \|_1
\end{equation}
For descriptive convenience, we introduce a weighting scalar $\beta\in(0, 1)$ to combine $L_{MS-SSIM}$and $L_{l_1}$ into a single term $L^{l_1}_{MS-SSIM}$ we comined $L_{MS_SSIM}$ and $L_{l_1}$ into one term $L^{l_1}_{MS_SSIM}$, it can be described as
\begin{equation}
	\label{equ7}
	L^{l_1}_{MS-SSIM}=(1-\beta)L_{MS-SSIM}+\beta L_{l_1}
\end{equation}
\begin{equation}
	\label{equ8}
	l_M(O,I)=\frac{2\mu_O\mu_I+C_1}{\mu_O^2+\mu_I^2+C_1}
\end{equation}
\begin{equation}
	\label{equ9}
	cs_j(O,I)=\frac{2\sigma_{OI}+C_2}{\sigma_O^2+\sigma_I^2+C_2}
\end{equation}
 where $\mu_I$, $\sigma_I^2$ and $\sigma_{OI}$ represent the mean of $I$, the variance of $I$ and the covariance of $I$ and $O$, respectively. $j=\left\{1,2,\cdots, M\right\}$, $C_1$, $C_2$ is small constant.\\
\hspace*{1em}Inspired by\cite{zhou2022unified}, \cite{xu2020deep}, the gradient loss function can force fused images to obtain richer texture information ,and it is defined as
\begin{equation}
	\label{equ10}
	L_{grad}=\left\|\nabla O-\nabla I\right\|_1
\end{equation}
Combining Eq.(\ref{equ1}), Eq.(\ref{equ2}), Eq.(\ref{equ7}) and  Eq.(\ref{equ10}), the total loss function can be expressed as
\begin{equation}
	\label{equ11}
    L_{total}=\alpha_1L_{pixel}+\alpha_2L_{ssim}+\alpha_3L^{l_1}_{MS-SSIM}+\alpha_4L_{grad}
\end{equation}
where $\alpha_1, \quad \alpha_2, \quad \alpha_3, \quad \alpha_4$ are the positive tuning parameters.
\subsection{Fusion Strategy}
The fusion layer is essential for generating fused images, and we implement three distinct fusion strategies: (1) weighted average method, (2) $L_1$-norm method, and (3) the mean operator method.\\
\hspace*{1em}After passing the input images through the encoder, feature maps are obtained. To create fused maps, using a special fusion strategy to perform weighted fusion of feature maps from different modalities. Subsequently, the fused maps are input into the decoder to generate a high-resolution fused image.\\
\textbf{1) Weight average method}\\
As a simple fusion strategy \cite{ram2017deepfuse}, it generally don't need complex design process and retain robust fusion perform.  
\hspace*{1em}Where $f^{m}_{i}(x,y)$ denotes the feature map obtained by encoder from input images, $f^{m}_{fused}(x,y)$ denotes the fused featured maps. The weight average strategy formulated by Eq.(\ref{equ12})
\begin{equation}
\label{equ12}
    f_{fused}^{m}(x,y)=\sum_{i=1}^{2}\frac{f_i^m(x,y)}{2}
\end{equation}
\textbf{2) $L_1$-norm method}\\
\hspace*{1em}Inspired by \cite{li2019infrared}, we use $L_1$-norm operator as a measure activity, and obtain the weighted maps $\omega_i$, which is computed by
\begin{equation}
    \label{equ13}
    \omega_i=\frac{\sum_{j=1}^m\Vert f_i^j(x,y)\Vert_1}{\sum_{i=1}^k\sum_{j=1}^m\Vert f_i^j(x,y)\Vert_1}
\end{equation}
\hspace*{1em}Where $f_i^{1:m}(x,y)$ are the feature maps extracted by the encoder, $m$ and $k$ denote the number of the feature maps and input images,respectively.
Finally, the fused maps can be obtained as
\begin{equation}
    \label{equa14}
    f_{fused}^{1:M}(x,y)=\sum_{i=1}^k\omega_i\times f_i^{1:M}(x,y)
\end{equation}
\textbf{3) Mean filter operator method:}\\
\hspace*{1em}We use the mean filter operator to process the feature map extracted by the encoder, the adding weights of feature maps can be calculated by
\begin{equation}
    \label{equ15}
    \omega_i(x,y)=\frac{\varphi(\Vert f_i(x,y\Vert_1)}{\sum_{i=1}^k\varphi(\Vert f_i(x,y)\Vert_1)}
\end{equation}
where $\varphi(\cdot)$ is a $3\times3$ mean filter. Consequently, the fused feature maps can be expressed as
\begin{equation}
    \label{equ16}
    f_{fused}^{1:M}=\sum_{i=1}^kw_i\times f_i^{1:M}(x,y)
\end{equation}
\section{ EXPERIMENTS AND DISCUSSIONS}
\hspace*{1em} In this section, we describe the experimental settings and environment. Subsequently, we compare the performance of our proposed model with other state-of-the-art (SOTA) models, including two traditional methods, namely GTF\cite{ma2016infrared} and MDLatLRR\cite{li2020mdlatlrr}, as well as four deep learning-based methods, specifically FusionGAN\cite{ma2019fusiongan}, Densefuse\cite{li2018densefuse}, DIDfuse\cite{zhao2020didfuse}, and U2Fusion\cite{xu2020u2fusion}. All experiments were conducted using PyTorch on a computer equipped with an Intel Xeon CPU@2.2GHz and an RTX 3090 GPU.
\subsection{Datasets and Preparation}
 During the training stage, we randomly selected 180 pairs of images from the dataset \cite{xu2020fusiondn} to train our proposed model. Before training, all images were converted to grayscale, and a center-crop of $224 \times 224$ was applied to the input images. The pixel intensity of the input images was normalized to a range between $-1$ and $1$. For the training phase, we optimized the loss function using the Adam solver. In the testing phase, we employed two datasets, TNO \cite{toet2012progress} and FLIR (available at https://github.com/jiayi-ma/RoadScene), to evaluate the efficiency and performance of the proposed model.
\subsection{Training Settings and Fusion Metrics}
\hspace*{1em}Our training parameters were established with a batch image size of 12 and 160 training iterations. The hyperparameters within the loss function were empirically determined: $\alpha_1=1$, $\alpha_2=1$, $\alpha_3=2$, $\alpha_4=0.005$, $\alpha_5=10$, and $\beta=0.0025$.
Visual assessment of fusion performance can be intricate; hence, we utilized quantitative fusion metrics for an unbiased evaluation. In our study, we adopted eight metrics: entropy (EN) \cite{roberts2008assessment}, denoting the image's informational content; mutual information (MI) \cite{qu2002information}, gauging the similarity between input image pairs and the fused output; SSIM \cite{wang2002universal}, determining the structural resemblance between the source and fused images; average gradient (AG) \cite{wu2005remote}, indicative of the image's detail representation and its clarity; standard deviation (SD) \cite{rao1997fibre}, portraying the image's distribution and contrast; spatial frequency (SF) \cite{eskicioglu1995image}, evaluating the gradient distribution and the image's detail and texture; $Q_{abf}$ 
 \cite{xydeas2000objective}, representing the quality of visual data; and visual information fidelity (VIF) \cite{han2013new}, assessing the visual data fidelity. For these metrics, higher values signify superior performance.
 \subsection{Experiments on fusion strategy}
In this section, we evaluate the performance of three fusion strategies using 40 randomly selected image pairs from both the TNO and FLIR datasets. The assessment relies on the average metrics derived from these image pairs. Table \ref{table_fuse} displays the performance outcomes based on nine evaluation metrics. Notably, the $L_1$-norm fusion strategy achieved four best values in the FLIR dataset and three best values in the TNO dataset. Therefore, subsequent experiments will adopt the $L_1$-norm fusion strategy.
\begin{table}[tbp]
	\centering
 	\caption{\centering{Results of validation set for choosing the addition strategy}}
	\begin{tabular}{cccc}
		\toprule
		\multicolumn{4}{c}{\textbf{Dataset: FLIR Dataset
		}}\\
		Method&\makebox[1.4cm][c]{Summation}&\makebox[1.4cm][c]{Average}&\makebox[1.4cm][c]{$L_1$-norm}\\
		\midrule
		EN	&	{7.3829} 	&	\textbf{7.3934} 	&	7.3869 	\\
            AG	&	\textbf{5.9153}	&	5.7364	&	5.8207 \\
		MI	&	2.7693 	&	2.8365 	&	\textbf{2.8520}	\\
		SD	&	51.3609 	&	\textbf{51.8447}	&	51.5468	\\
		SF	&	\textbf{14.9921} &	14.6948	&	14.8068	\\
            $Q_{abf}$	&	0.4661 	&	0.4598 		&	\textbf{0.4732}           \\
            SSIM	&	0.9680	&	0.9703	&	\textbf{0.9745} 
            \\
		VIF	&0.5782 	&0.5874 	&	\textbf{0.5906} 	\\
		SCD	&	\textbf{1.6975}	&	1.6924	&	1.6892	\\
          \midrule
		\multicolumn{4}{c}{\textbf{Dataset: TNO Dataset}}\\
		Method&\makebox[1.5cm][c]{Summation}&\makebox[1.5cm][c]{Average}&\makebox[1.5cm][c]{$L_1$-norm}\\	
		\midrule	
		EN	&	7.3431	&	\textbf{7.3762} 	&	7.3654 	\\
            AG	&	\textbf{5.9753}	&	5.8069	&	5.8853 
  \\
		MI	&	2.2406	&	\textbf{2.3294} 	&	2.3226	\\
		SD	&	47.0399	&	\textbf{48.5152}	&	47.7786	\\
		SF	&	\textbf{14.8042} &	14.5458	&	14.6055	\\
            $Q_{abf}$	&	0.3718 	&	\textbf{0.3872} 		&	0.3810           \\
            SSIM	&	0.8396	&	0.8494	&	\textbf{0.8538}
            \\
		VIF	&0.6535 	&0.6590 	&	\textbf{0.6683} 	\\
		SCD	&	1.6922	&	1.6871	&	\textbf{1.6987}	\\
		\bottomrule
	\end{tabular}
\label{table_fuse}
\end{table}
 \subsection{Experimental Result and Analysis}
\begin{figure}[htbp]
\centering
\includegraphics[width=1.0\linewidth]{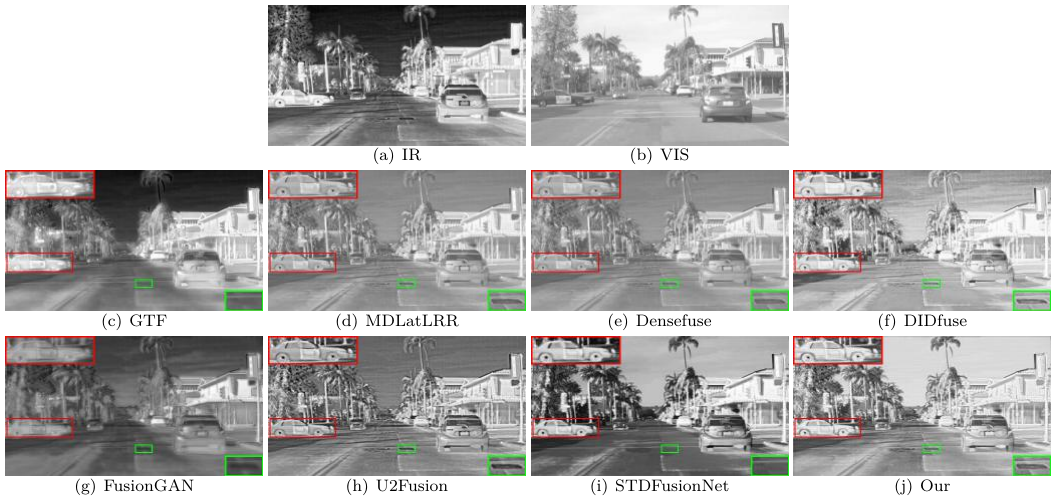}
\caption{Qualitative comparison of AMFusionNet with seven state-of-the-art methods on  \textit{FLIR\_04424}. For a clear comparison, we select a salient region(\emph{i.e.}, the red box and green color box) in each image and zoom in on it using the same color box, respectively. From top to bottom (from left to right): infrared and infrared image pair, fusion results of GTF \cite{ma2016infrared}, MDLatLRR \cite{li2020mdlatlrr}, STDFusionNet \cite{ma2021stdfusionnet}, FusionGAN \cite{ma2019fusiongan}, Densefuse \cite{li2018densefuse}, DIDfuse \cite{zhao2020didfuse}, U2Fusion \cite{xu2020u2fusion}, our AMFusionNet.}\label{figss}%文中引用该图片代号
\label{flrfig2}
\end{figure}

\begin{figure}[htbp]
\centering
\includegraphics[width=1.0\linewidth]{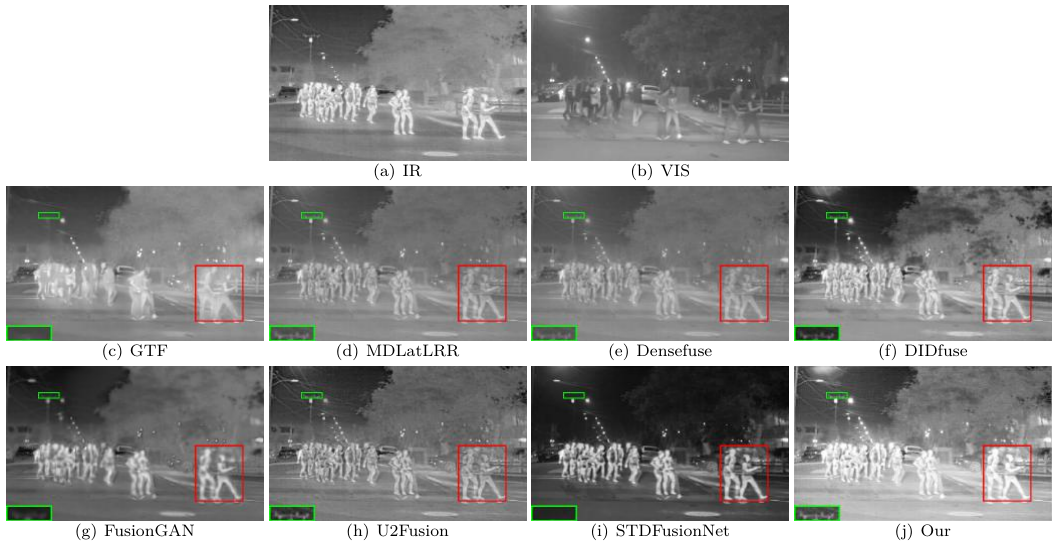}
\caption{Qualitative comparison of AMFusionNet with seven state-of-the-art methods on  \emph{FLIR\_07620}. For a clear comparison, we select a salient region(\emph{i.e.}, the red box and green color box) in each image and zoom in on it using the same color box, respectively. From top to bottom (from left to right): infrared and infrared image pair, fusion results of GTF \cite{ma2016infrared}, MDLatLRR \cite{li2020mdlatlrr}, STDFusionNet \cite{ma2021stdfusionnet}, FusionGAN \cite{ma2019fusiongan}, Densefuse \cite{li2018densefuse}, DIDfuse \cite{zhao2020didfuse}, U2Fusion \cite{xu2020u2fusion}, our AMFusionNet.} \label{figss}%文中引用该图片代号
\label{flrfig3}
\end{figure}

\begin{figure}[htbp]
\centering
\includegraphics[width=1.0\linewidth]{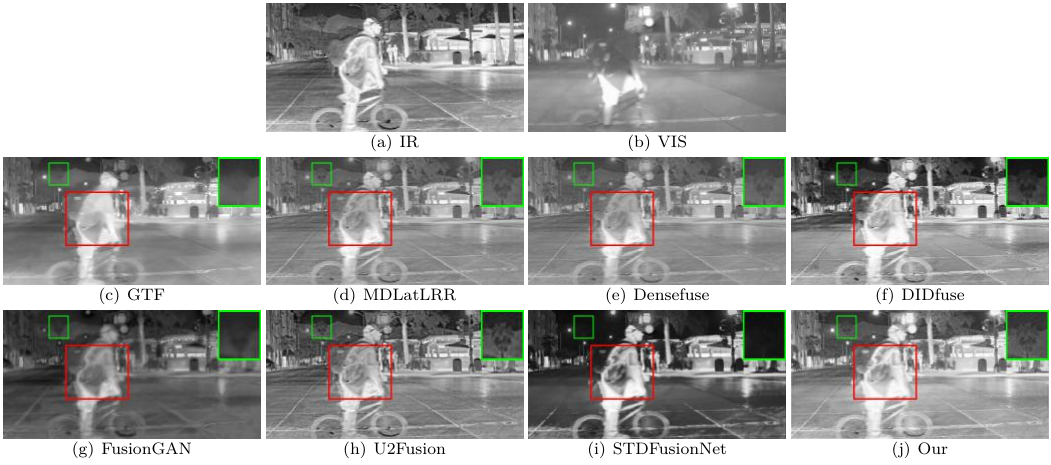}
\caption{Qualitative comparison of AMFusionNet with seven state-of-the-art methods on  \textit{FLIR\_09488}. For a clear comparison, we select a salient region(\emph{i.e.}, the red box and green color box) in each image and zoom in on it using the same color box, respectively. From top to bottom (from left to right): infrared and infrared image pair, fusion results of GTF \cite{ma2016infrared}, MDLatLRR \cite{li2020mdlatlrr}, STDFusionNet \cite{ma2021stdfusionnet}, FusionGAN \cite{ma2019fusiongan}, Densefuse \cite{li2018densefuse}, DIDfuse \cite{zhao2020didfuse}, U2Fusion \cite{xu2020u2fusion}, our AMFusionNet.}\label{figss}%文中引用该图片代号
\label{flrfig4}
\end{figure}

 In this section, we evaluate the performance of our proposed model compared to other state-of-the-art methods. We conduct both qualitative and quantitative evaluations using publicly available TNO and FLIR datasets. The qualitative comparison results on the FLIR dataset are shown in Fig.\ref{flrfig2}-\ref{flrfig4}

 1) Qualitative Results: For a direct comparison of the image fusion efficacy between our proposed algorithm and existing methods, we selected three representative pairs of source images from the FLIR dataset. Fig.\ref{flrfig2}-\ref{flrfig4} display the fusion results obtained using various algorithms. \\
\hspace*{1em}In Fig.\ref{flrfig2}-\ref{flrfig4}, we identify two regions for comparative analysis: a background region and a salient target region. Fig.\ref{flrfig2} reveals that FusionGAN omits crucial thermal emission target details, hence missing the infrared information of the car. While STDFusionNet, GTF, and U2Fusion emphasize the salient target, they inadequately capture significant background context, like the manhole cover. Conversely, our AMFusionNet adeptly differentiates the target from its backdrop, also ensuring comprehensive retention of the visible image's background texture. In Fig.\ref{flrfig3}, all algorithms consistently maintain thermal details, facilitating a clear distinction between the target and background. However, when it comes to conserving background specifics, our method excels. It retains both the infrared image's thermal details and the visible image's texture, as highlighted in the magnified red box. This enhanced performance underscores the proficiency of our technique.\\
 \hspace*{1em}In Fig.\ref{flrfig4}, FusionGAN and MDLatLRR are observed to degrade the thermal radiation details of the infrared salient target due to the visible light image's limitations under low-light conditions. While STDFusionNet and GTF manage to retain a majority of the infrared data, they falter in preserving certain background details, evident in the blurred boundary of the tree. U2Fusion effectively conserves both infrared target and background texture details. However, when compared to our method, it retains lesser infrared information.\\
 \hspace*{1em}Upon comparison, it is evident that our proposed algorithm produces fused images with improved target brightness, more distinct edge contours, and superior retention of detailed information. Examples include the manhole cover in Fig.\ref{flrfig2} and the man riding a bicycle in Fig.\ref{flrfig4}.\\
\begin{figure*}[htbp]
    \centering
    \includegraphics[width=1.0\linewidth]{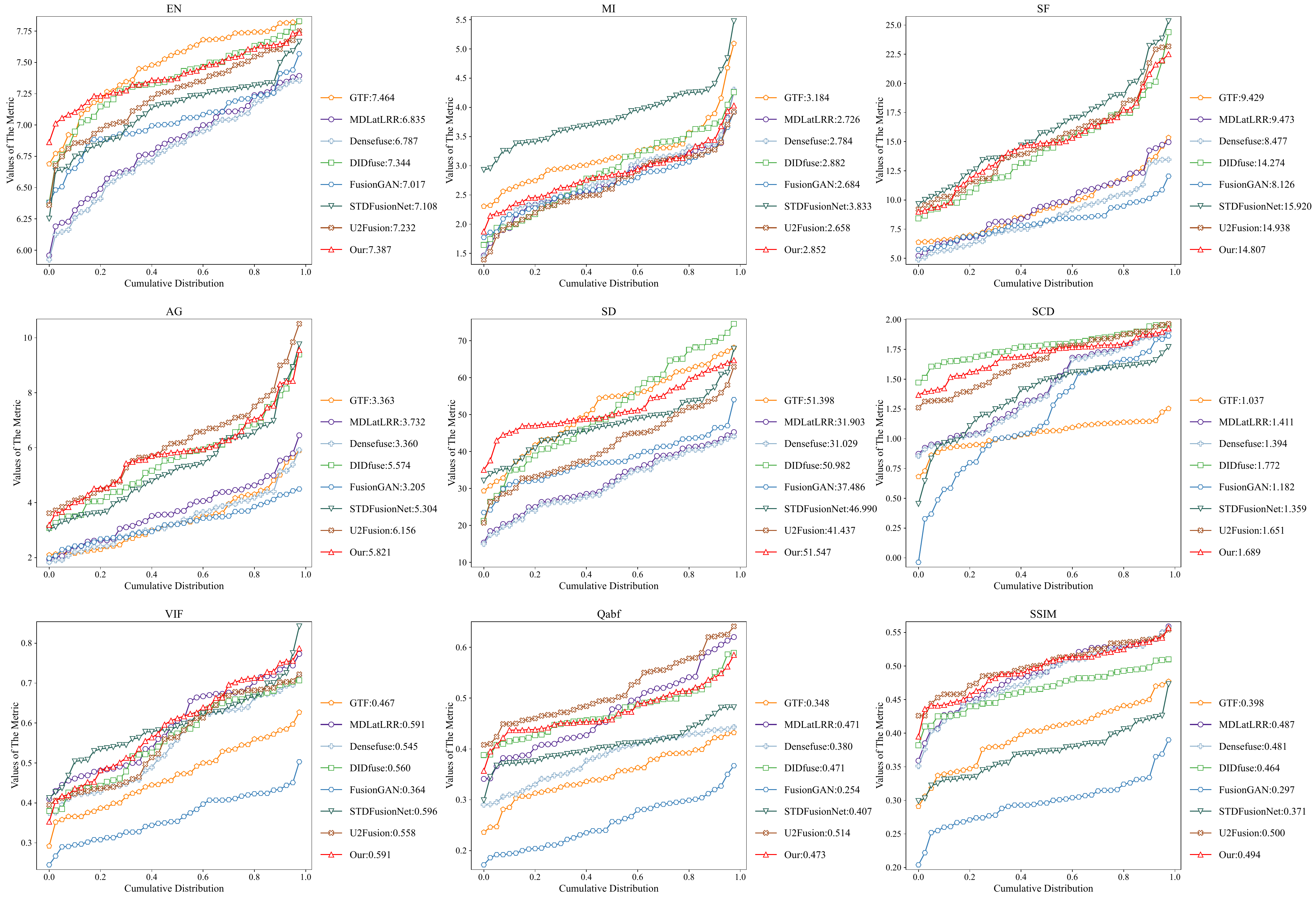}
    \caption{ The subjective comparative results of eight evaluation metrics for the TNO dataset, accompanied by the corresponding average values of different fusion methods. Notably, our proposed AMFusionNet is indicated by a red dotted line.}
    \label{flrfig5}
\end{figure*}
 2) Quantitative Results: We assessed the performance using nine widely recognized quantitative metrics on 40 image pairs from the FLIR dataset, as depicted in Fig.\ref{flrfig5} and Table \ref{table2}. Our algorithm secures the top position in the SCD evaluation metric compared to seven other cutting-edge algorithms. Additionally, it achieves the best score in one metric and ranks second in eight metrics, including EN, AG, MI, SD, SF, $Q_{ABF}$, and VIF. Despite these second-place rankings, our algorithm consistently displays competitive performance across these metrics, underscoring its capability to generate superior quality fused images.\\
 \begin{table*}[htbp]
	\renewcommand\arraystretch{1.5}
	%\renewcommand\tabcolsep{3.5pt}
	%% increase table row spacing, adjust to taste
	%\renewcommand{\arraystretch}{1.3}
	% if using an array.sty, it might be a good idea to tweak the value of
	% \extrarowheight as needed to properly center the text within the cells
	\caption{\centering{QUANTITATIVE COMPARISONS OF THE FOUR METRICS, i.e., EN, AG, MI, SD, SF, $Q_{abf}$, SSIM, VIF, SCD, ON 40 IMAGE PAIRS FROM THE TNO  DATASETS. \textcolor{red}{RED} INDICATES THE BEST RESULT AND \textcolor{blue}{BLUE} INDICATES THE SECOND BEST RESULT.}}
	\label{table2}
	\centering
	%% Some packages, such as MDW tools, offer better commands for making tables
	%% than the plain LaTeX2e tabular which is used here.
    \begin{tabular}{l c c c c c c c c c}
		\hline
		Methods  & EN & AG & MI & SD & SF & Q$_{abf}$ &SSIM & VIF & SCD    \\
		\hline
		GTF  & \textcolor{red}{7.4641} & 3.3632 & \textcolor{blue}{3.1839} &\textcolor{blue}{51.3977}&9.4291 &0.3480& 0.7914 &0.4673 &1.0373 \\
		\hline 
            MDLatLRR  & 6.8351&3.7318 &2.7263 & 31.9031 &9.473 &0.4707 &{0.9672} & 0.5906 &1.411 \\
            \hline
        STDFusionNet & 7.1080 &{5.3042} &\textcolor{red}{3.8334} &{46.9905} &\textcolor{red}{15.9198} &0.4066 & 0.7367 &\textcolor{red}{0.596} &1.3588\\
          \hline
        U2Fusion  & 7.2324 & \textcolor{red}{6.1577} & 2.6577 & 41.4372 &\textcolor{blue}{14.9383} &\textcolor{red}{0.5145} &\textcolor{red}{0.9941} &0.5584 &1.6513\\
          \hline
        FusionGAN  &7.0167 & 3.2046& 2.6836 & 37.4861& 8.1258 &0.2536 &0.5951&0.3638 &1.1815 \\
          \hline
        Densefuse  &6.7865&3.3599&2.7839&31.0294&8.4774&0.3801&{0.9569}&0.5446&1.3938\\
          \hline
        DIDfuse  & 7.3441 &{5.5742}& 2.8822 & 50.9818 &14.2743& {0.4713}&0.9166&0.5602&\textcolor{red}{1.7719} \\
          \hline
        Our  & \textcolor{blue}{7.3841}&\textcolor{blue}{5.8207}&{2.8520}&\textcolor{red}{51.5468}&{14.8068}&\textcolor{blue}{0.4732}&\textcolor{blue}{0.9745}&\textcolor{blue}{0.5906}&\textcolor{blue}{1.6892}\\
          % \hline
          % Our\_{add}  & 7.3830& \textcolor{blue}{5.9153}&14.7659&51.36096& \textcolor{blue}{14.9706} &{0.4661}&{0.6817}&0.7463&\textcolor{red}{1.6216}\\
          % \hline
          % Our\_{channel}  & \textcolor{blue}{7.3934}&\textcolor{green}{5.7364}&\textcolor{blue}{14.7869}&\textcolor{blue}{51.8447} &14.6739&0.4598&{0.6863}&\textcolor{green}{0.7586}&1.5933\\
          \hline
    \end{tabular}
\end{table*}
\hspace*{1em}As depicted in Fig.\ref{flrfig2}-\ref{flrfig4} and detailed in Table \ref{table2}, the efficacy of the algorithm can be both subjectively gauged through visual inspection and objectively measured using nine distinct metrics. Recognizing the diverse strengths of each algorithm across these metrics, we propose a new evaluation metric, termed the Normalized Evaluation Index, denoted as:
\begin{equation}
\label{equ17}
    \xi(i)=\sum_{j=1}^{N}\frac{f_{j}(x_i)}{ \mathop{\max}_{x_i}f_j(x_i)}
\end{equation}
\hspace*{1em}The sum of the normalized values for each algorithm across these nine metrics is presented in the table below:
 \begin{table*}[htbp]
	\renewcommand\arraystretch{1.5}
	%\renewcommand\tabcolsep{3.5pt}
	%% increase table row spacing, adjust to taste
	%\renewcommand{\arraystretch}{1.3}
	% if using an array.sty, it might be a good idea to tweak the value of
	% \extrarowheight as needed to properly center the text within the cells
	\caption{\centering{QUANTITATIVE COMPARISONS OF THE FOUR METRICS, i.e., EN, AG, MI, SD, SF, $Q_{abf}$, SSIM, VIF, SCD, ON 40 IMAGE PAIRS FROM THE TNO  DATASETS. \textcolor{red}{RED} INDICATES THE BEST RESULT.}}
	\label{table3}
	\centering
	%% Some packages, such as MDW tools, offer better commands for making tables
	%% than the plain LaTeX2e tabular which is used here.
  \resizebox{0.9\textwidth}{!}{
    \begin{tabular}{l c c c c c c c c c}
		\hline
		 Method& GTF & MDLatLRR &STDFusionNet &U2Fusion & FusionGAN &Densefuse&DIDfuse & $Our$  \\
        Normalized Values &6.8079&7.1220&8.0235&8.2730&5.7668&6.7175&{8.3046}&\textcolor{red}{8.4532} \\
		\hline
    \end{tabular}}
\end{table*}
\hspace*{1em}Considering the aggregated normalized values across these nine metrics, our algorithm achieved the highest value, surpassing the third-ranked algorithm by 2.18\%. This underscores our algorithm's superior overall performance.\\
\subsection{Generalization Analysis}
\begin{figure}[htbp]
\centering
\includegraphics[width=1.0\linewidth]{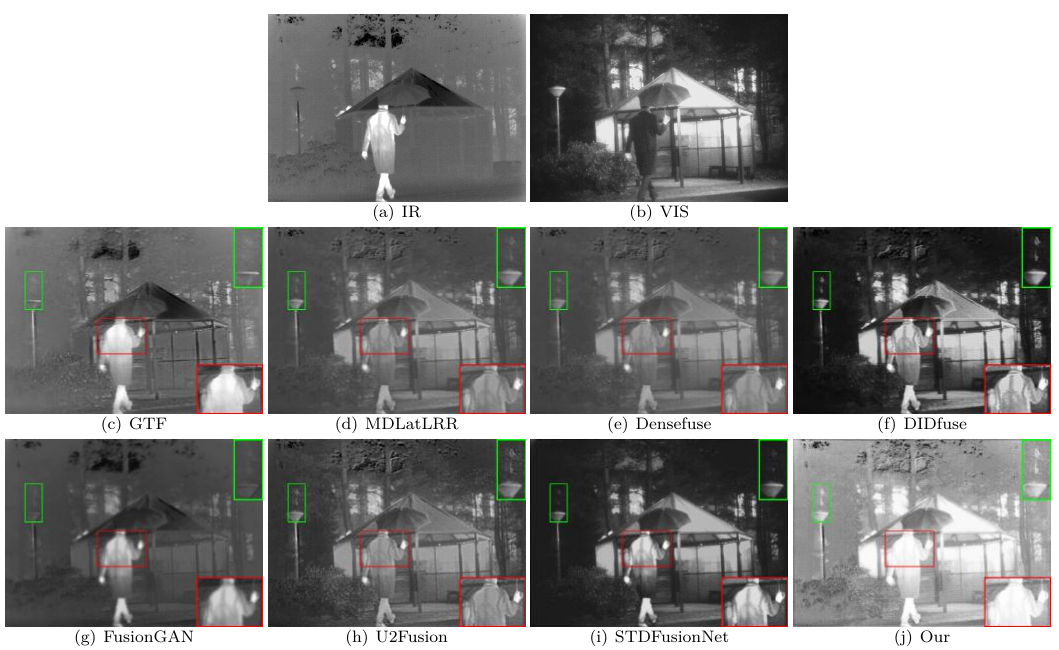}
\caption{Qualitative comparison of AMFusionNet with seven state-of-the-art methods on  \textit{soldier\_behind\_smoke}. For a clear comparison, we select a salient region(\emph{i.e.}, the red box and green color box) in each image and zoom in on it using the same color box, respectively. From top to bottom (from left to right): visible and infrared image pair, fusion results of GTF \cite{ma2016infrared}, MDLatLRR 
 \cite{li2020mdlatlrr}, STDFusionNet \cite{ma2021stdfusionnet}, FusionGAN\cite{ma2019fusiongan}, Densefuse 
 \cite{li2018densefuse}, DIDfuse \cite{zhao2020didfuse}, U2Fusion \cite{xu2020u2fusion}, our AMFusionNet.}
\label{tnofig6}
\end{figure}

\begin{figure}[htbp]
\centering
\includegraphics[width=1.0\linewidth]{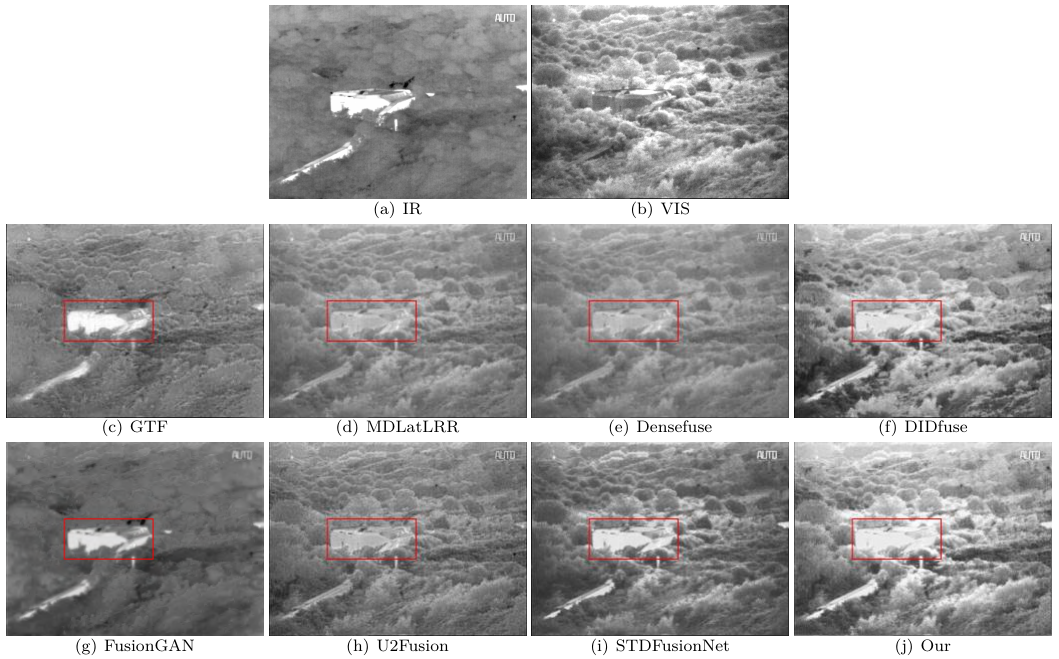}
\caption{Qualitative comparison of AMFusionNet with seven state-of-the-art methods on  \textit{Kaptein\_1654}. For a clear comparison, we select a salient region(\emph{i.e.}, the red box and green color box) in each image and zoom in on it using the same color box, respectively. From top to bottom (from left to right): visible and infrared image pair, fusion results of GTF \cite{ma2016infrared}, MDLatLRR \cite{li2020mdlatlrr}, STDFusionNet \cite{ma2021stdfusionnet}, FusionGAN \cite{ma2019fusiongan}, Densefuse \cite{li2018densefuse}, DIDfuse \cite{zhao2020didfuse}, U2Fusion \cite{xu2020u2fusion}, our AMFusionNet.}
\label{tnofig7}
\end{figure}

\begin{figure}[htbp]
\centering
\includegraphics[width=1.0\linewidth]{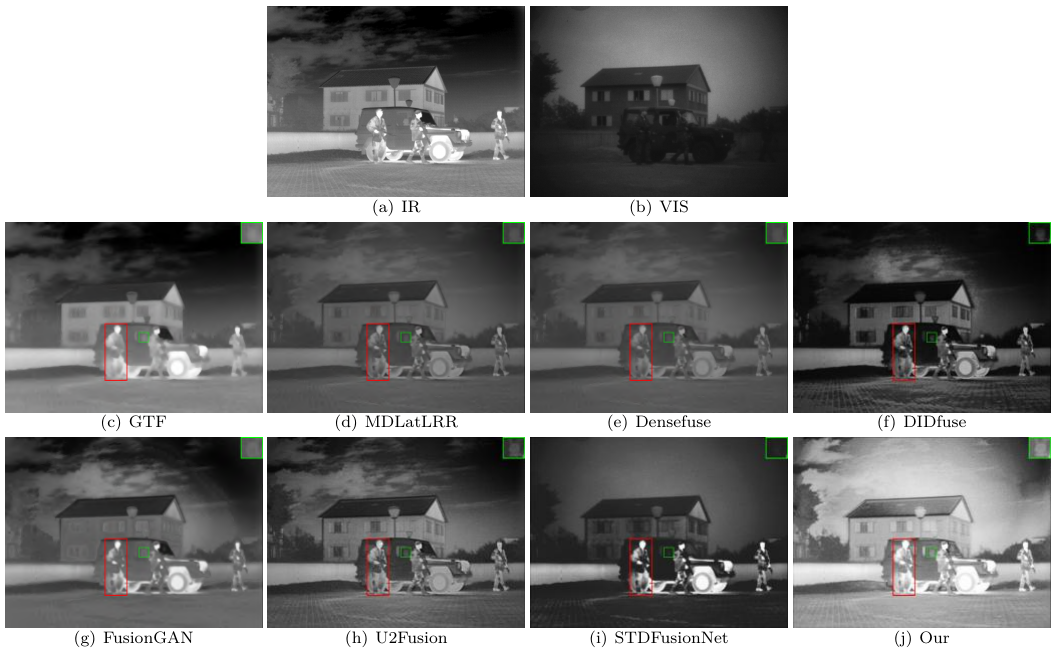}
\caption{Qualitative comparison of AMFusionNet with seven state-of-the-art methods on  \emph{Jeep}. For a clear comparison, we select a salient region(\emph{i.e.}, the red box and green box) in each image and zoom in it use same color box, respectively. From top to bottom (from left to right): visible and infrared image pair, fusion results of GTF\cite{ma2016infrared}, MDLatLRR\cite{li2020mdlatlrr}, STDFusionNet\cite{ma2021stdfusionnet}, FusionGAN\cite{ma2019fusiongan}, Densefuse\cite{li2018densefuse}, DIDfuse\cite{zhao2020didfuse}, U2Fusion\cite{xu2020u2fusion}, our AMFusionNet.}
\label{tnofig8}
\end{figure}

\begin{figure*}
    \centering
     \includegraphics[width=1.0\linewidth]{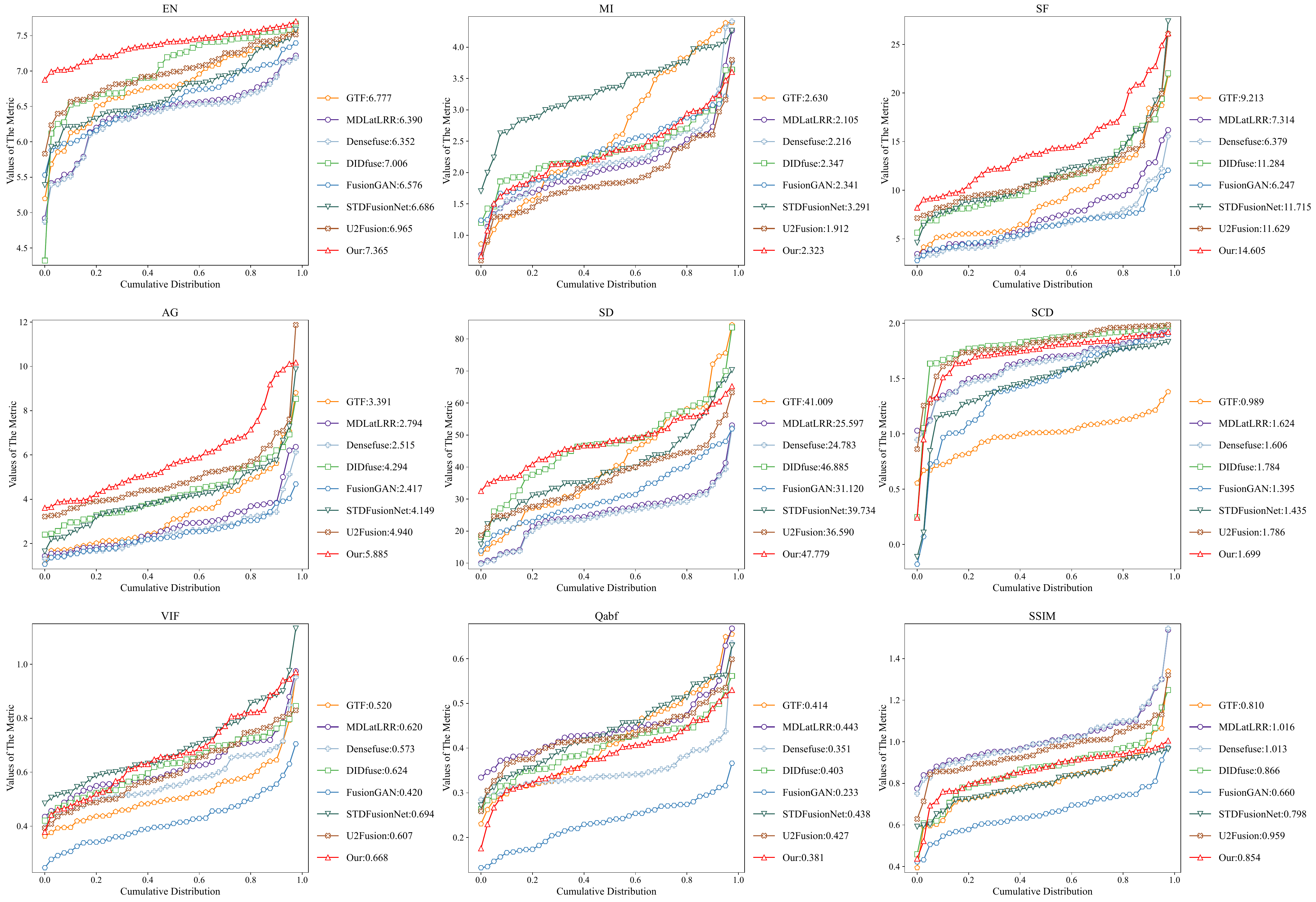}
    \caption{ The objective comparative results of eight evaluation metrics for the TNO dataset, accompanied by the corresponding average values of different fusion methods. Notably, our proposed AMFusionNet is indicated by a red dotted line.}
    \label{tnofig9}
\end{figure*}

Evaluating the generalization capability of a deep learning model is crucial for determining its overall efficacy. To assess the generalization performance of our AMFusionNet model, we test it on image pairs from the dataset, with the model having been trained on the TNO dataset.\\
\hspace*{1em}1)Qualitative Results: Figures \ref{tnofig6}-\ref{tnofig8} display the fusion results from different methods. Our algorithm effectively retains the thermal information from infrared images and accentuates the boundaries of prominent targets in the fused imagery. Compared to other methods, our approach excels in maintaining more background data, achieving superior contrast, elucidating additional details, and delivering more discernible targets. For instance, in Fig.\ref{tnofig6}, MDLatLRR's fusion result omits the infrared target's radiation information. While the fusion outcomes from DIDfuse and STDFusionNet can identify the infrared prominent target, their overall fused structure undergoes smoothing effects that lead to environmental distortions, such as the underemphasized radiation details of streetlights. The fusion outputs of Densefuse, GTF, and U2Fusion are notably affected by noise. In contrast, our algorithm's fusion results allow for easy differentiation of infrared prominent targets with well-defined boundaries. \\
\hspace*{1em}2) To objectively assess the performance of various fusion algorithms, we test 40 image pairs from the TNO dataset. The fusion outcomes are presented in Fig.\ref{tnofig9} and Table \ref{table4}, indicating that our algorithm exhibits strong performance across these nine metrics. Specifically, for the TNO dataset, our method achieves the highest scores in EN, AG, MI, SD, SF, and SCD, and ranks second in VIF. A superior EN score suggests our fused images encapsulate richer information compared to alternatives. Elevated AG, SD, and SF scores signify results that are not only more informative but also visually compelling, adeptly retaining detail, contrast, and texture from the source imagery. Similarly, the top-ranking MI and SCD values underscore our technique's proficiency in transferring maximal information from the source to the fused images.\\
\hspace*{1em}In conclusion, the images fused using our approach are superior for visual perception, offering enhanced contrast, distinct salient targets, and negligible texture degradation. Objectively and subjectively, these attributes underscore the robust generalization capabilities of our method


 \begin{table*}[htbp]
	\renewcommand\arraystretch{1.5}
	%\renewcommand\tabcolsep{3.5pt}
	%% increase table row spacing, adjust to taste
	%\renewcommand{\arraystretch}{1.3}
	% if using array.sty, it might be a good idea to tweak the value of
	% \extrarowheight as needed to properly center the text within the cells
	\caption{\centering{Quantitative Comparisons Of The Nine Metrics, \textit{i.e.}, EN, AG, MI, SD, SF, $Q_{abf}$, SSIM, VIF, SCD, On 40 Image Pairs From The TNO Datasets. \textcolor{red}{Red} Indicates The Best Result \textcolor{blue}{BLUE} Indicates The Second Best Result.}}
	\label{table4}
	\centering
	%% Some packages, such as MDW tools, offer better commands for making tables
	%% than the plain LaTeX2e tabular which is used here.
    \begin{tabular}{l c c c c c c c c c}
		\hline
		Methods  & EN & AG & MI & SD & SF & Q$_{abf}$ &SSIM & VIF & SCD    \\
		\hline
		GTF  & 6.7772 & 3.3911 &\textcolor{blue}{2.6304} & 41.0091 & 9.2126 &{0.4136} & {0.8099} &0.5202&0.9888 \\
		\hline 
            MDLatLRR  & 6.3904 & 2.7937 &2.1048 & 25.5972 & 7.3137 &\textcolor{red}{0.4427} &\textcolor{red}{1.0158}&0.6195 &1.6237 \\
            \hline
        STDFusionNet & 6.8700 &4.2318 &\textcolor{red}{3.2912} & {39.734} & 11.7147 &\textcolor{blue}{0.4376} & 0.7982 &\textcolor{red}{0.694} &1.4352 \\
          \hline
        U2Fusion  & 6.9655 & \textcolor{blue}{4.94} & 1.1924 & 36.5903 & \textcolor{blue}{11.6162} &0.4266 & 0.9588 &0.6066 &\textcolor{red}{1.7856} \\
          \hline
        FusionGAN  &6.5761 & 2.41691& 2.341 & 31.1199 & 6.2466 &0.2328 & 0.6603 &0.4201 &1.1.3955 \\
          \hline
        Densefuse  &6.3518  & 2.5148 & 2.216 &24.7829 & 6.3794 &0.3506& \textcolor{blue}{1.0127} &0.5727 &1.6056\\
          \hline
        DIDfuse  & \textcolor{blue}{7.0061} & {4.2942}& 2.3468 &\textcolor{blue} {46.8854} &{11.2839}&0.4027 &0.8658&0.6235 &\textcolor{red}{1.7837} \\
          \hline
       Our &\textcolor{red}{7.3654} &\textcolor{red}{5.8853} & {2.3226} & \textcolor{red}{47.7786} &\textcolor{red}{14.6055} &{0.3810} &0.8538&\textcolor{blue}{0.6683}&{1.6987}\\
      \hline
      %  Our\_add  &\textcolor{green}{7.3432} & \textcolor{blue}{5.9753} & \textcolor{green}{14.6864} & \textcolor{green}{47.0399}&\textcolor{red}{14.7878} & 0.3718& 0.5720 &{0.9057}&\textcolor{blue}{1.6680}\\
      % \hline
      %   Our\_channel  & \textcolor{blue}{7.3762} & {5.8069} & \textcolor{red}{14.7524} & \textcolor{red}{48.5152} & \textcolor{green}{14.5297}&{0.3872} & 0.5811&\textcolor{green}{0.9119}&\textcolor{blue}{1.6574}\\
      %     \hline
    \end{tabular}
\end{table*}
\hspace*{1em}According to Eq.(\ref{equ17}), we can calculate the normalized evaluation metrics for the fused images generated by AMFUSIONnet and the other seven algorithms, as show in Table \ref{biao5}
 \begin{table*}[htbp]
	\renewcommand\arraystretch{1.5}
	\caption{\centering{Quantitative Comparisons Of The Nine Metrics, \textit{i.e.}, EN, AG, MI, SD, SF, $Q_{abf}$, SSIM, VIF, SCD, On 40 Image Pairs From The TNO Datasets. \textcolor{red}{Red} Indicates The Best Result.}}
	\label{table4}
	\centering
	%% Some packages, such as MDW tools, offer better commands for making tables
	%% than the plain LaTeX2e tabular which is used here.
 \resizebox{0.9\textwidth}{!}{
    \begin{tabular}{l c c c c c c c c c}
		\hline
            \label{biao5}
		 Method& GTF & MDLatLRR &STDFusionNet &U2Fusion & FusionGAN &Densefuse&DIDfuse & $Our$    \\
        Normalized Values &6.8195&6.8203&7.8245&7.7098&5.6566&6.4318&{7.8071}&\textcolor{red}{8.3211}\\
		\hline
    \end{tabular}}
\end{table*}

\subsection{Ablation Experiment}
\begin{figure}[htbp]
\centering
\includegraphics[width=1.0\linewidth]{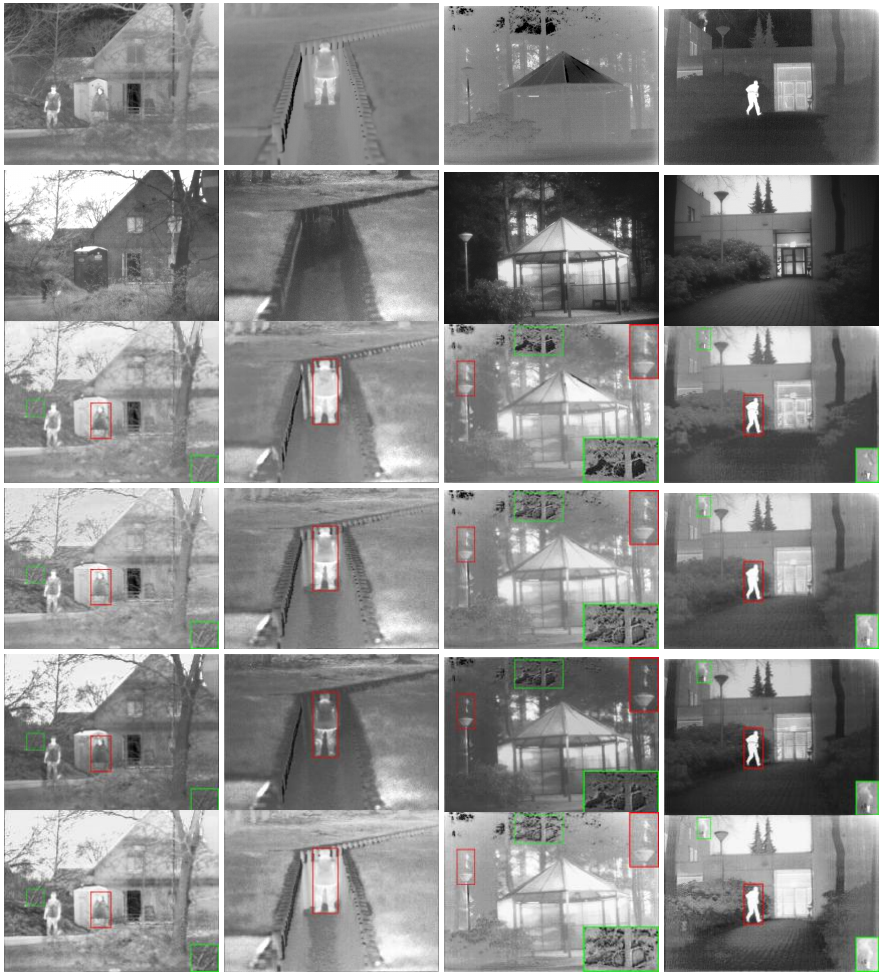}
\caption{Qualitative comparison of AMFusionNet with AMFusion-SC, AMFusion-MSSSIM and PSCNet on  \textit{men\_in\_front\_of\_house}, \textit{Kaptein\_1123}, \textit{Kaptein\_19} and \textit{solid\_inbranch}. For a clear comparison, we select a salient region(\emph{i.e.}, the red box and green color box) in each image and zoom in on it using the same color box, respectively. From top to bottom (from left to right): infrared and visible image pair, fusion results of AMFusionNet-SC, AMFusionNet-MSSSIM, PSCNet and AMFusionNet.}\label{fig9}
\end{figure}

\hspace*{1em}In this section, we assess the influence of attention mechanisms and MS-SSIM loss function on the performance of the AMFusionNet model. We explore three variations: AMFusionNet without the attention mechanism (denoted as AMFusionNet-SC), AMFusionNet employing a parallel form of the attention mechanism (labelled PSCNet), and AMFusionNet excluding the MS-SSIM loss function (referred to as AMFusionNet-MSSSIM).\\
\hspace*{1em}1)  Attention Mechanism Modules Analysis: The attention mechanism plays a pivotal role in feature extraction and fusion tasks \cite{wan2021lightweight, xiao2023mfmanet}. To evaluate its significance, we conducted an experiment in which we excluded the attention module but kept all other network components intact.\\
\hspace*{1em}The fusion outcomes of AMFusionNet and AMFusionNet-SC are depicted in Fig.\ref{fig9}. Significantly, the AMFusionNet not only highlights distinct targets within salient regions but also conserves the textural intricacies of background areas. Conversely, the AMFusionNet-SC network is less adept at preserving prominent infrared image details and at outlining sharp boundaries for salient targets compared to its AMFusionNet counterpart. Quantitative metrics further underscore the superior efficacy of AMFusionNet over AMFusionNet-SC. As detailed in Table \ref{table5}, AMFusionNet attains five leading metric scores and secures a runner-up position in one metric. \\
\hspace*{1em}2) MS-SSIM Loss Analysis: Incorporating the MS-SSIM loss function enhances the network's capability to retain texture information. This improvement stems from its method of comparing features between input and output images across multiple scales, sensitizing the network to minute image variations. As a result, the network is less prone to excessive image smoothing, ensuring the preservation of finer image details. In Fig.\ref{fig9}, qualitative comparison of AMFusionNet with  AMFusionNet-MSSSIM on \textit{men\_in\_front\_of\_house} and \textit{Kaptein\_19}, Compared to the fusion results of AMFUSIONNET, MM's fusion results contain more textures, such as the tree textures. From Table.\ref{table5}, it can be seen that AMFusionNet outperforms AMFusionNet-MSSSIM in four metrics: EN, AG, SD, and SF. EN measures the amount of information contained in a fused image; AG measures the richness of edge and texture information in the image; SF measures the complexity of the texture; and SD measures the contrast of the image. All three indicators intuitively reflect that AMFusionNet performs better than AMFusionNet-MSSSIM.\\
\hspace*{1em}3) Paralell Attention Mechanism Analysis: As previously discussed, the sequential attention mechanism is prone to information loss because each step's output depends on the output of the previous step. This shortfall arises from its inability to fuse the features from both the channel attention and spatial attention branches. Subjectively, as depicted in Figure \ref{fig9}, PSCNet falls short in retaining infrared thermal radiation data, yielding overly smoothed details in the fused image. In contrast, AMFusionNET adeptly preserves both information. Objectively, AMFusionNet fusion performance surpasses PSCNet in four metrics: EN, AG, SF, and SD, which validates our theoretical analysis.
\begin{table}[htbp]
	\renewcommand\arraystretch{1.8}
	%\renewcommand\tabcolsep{3.5pt}
	%% increase table row spacing, adjust to taste
	%\renewcommand{\arraystretch}{1.3}
	% if using an array.sty, it might be a good idea to tweak the value of
	% \extrarowheight as needed to properly center the text within the cells
	\caption{\centering{Quantitative Comparisons Of The Four Metrics, \textit{i.e.}, EN, AG, MI, SD, SF, \(Q_{abf}\), SSIM, VIF, SCD, On 40 Image Pairs From The TNO Datasets. \textcolor{red}{Red} Indicates The Best Result
}}
	\label{table5}
	\centering
 \resizebox{\linewidth}{!}{
    \begin{tabular}{c c c c c}
		\hline
		  & AMFusionNet-SC & AMFusionNet-MSSSIM&SCSCNet &AMFusionNet\\
		\hline
		EN  &7.2525  & 7.225&7.2579& \textcolor{red}{7.3654}\\
		\hline 
        AG &  4.5848 &4.8944 &4.8083& \textcolor{red}{5.8853}  \\
          \hline
        MI & 2.536  &2.239& \textcolor{red}{3.0903}  & {2.3226}  \\
          \hline
        SD  &45.8586 &42.0822 &44.7442  &\textcolor{red}{47.7786}\\
          \hline
        SF  &11.6601  &11.9025 &12.2281 &\textcolor{red}{14.6055}   \\
          \hline
        $Q_{abf}$  &{0.4253} &  0.3883& \textcolor{red}{0.4911}&  {0.3718}\\
          \hline
        SSIM  & {0.8892} & {0.9666}&\textcolor{red}{0.9959}& 0.8538  \\
          \hline
        VIF  & 0.6644 & \textcolor{red}{0.6879} &0.6015& {0.6683} \\
          \hline
          SCD  & 1.7023  &\textcolor{red}{1.7717}  &1.5722&{1.6987} \\
          \hline
             
    \end{tabular}}
\end{table}
% According to the Eq.(\ref{equ17}) and Table\ref{table5}, the calculated normalized metrics are as follows:\\
%  \begin{table}[htbp]
% 	\renewcommand\arraystretch{1.5}
% 	\caption{\centering{Normalized Metric Comparison. \textcolor{red}{RED} INDICATES THE BEST RESULT.}}
% 	\label{table6}
% 	\centering
% 	%% Some packages, such as MDW tools, offer better commands for making tables
% 	%% than the plain LaTeX2e tabular which is used here.
%   \resizebox{\linewidth}{!}{
%     \begin{tabular}{c c c c c}
% 		\hline
% 		 Method& AMFusionNet-SC &AMFusionNet-MSSSIM &SCSNet &AMFusionNet  \\
%    \hline
%         Normalized Values &8.3604&8.3912&\textcolor{red}{8.6088}\\
% 		\hline
%     \end{tabular}}
% \end{table}

\section{CONCLUSION}
\hspace*{1em}In this paper, we propose a novel infrared and visible image fusion network that leverages a cross-connected network structure with multiple convolutional kernels for feature extraction. This network employs a cross-connected structure utilizing multiple convolutional kernels for optimal feature extraction. In comparison to conventional single-kernel convolutional neural networks, our innovative approach broadens the network's receptive field. This minimizes information loss during the convolution phase, ensuring that the fused image captures a richer array of texture and thermal radiation details from the input images. Additionally, our design incorporates a parallel attention mechanism, bolstering the synergy between the channel attention and spatial attention networks. This refines the fused image's retention of salient and  details information. Concurrently, the integration of the multi-scale structural similarity (MS-SSIM) loss function proves pivotal. Optimizing against this function enables our fusion algorithm to adeptly conserve the structural nuances and pivotal features from the original images, culminating in a visually superior and content-rich fused output.  The test results on the FLIR and TNO datasets demonstrate that our algorithm outperforms existing state-of-the-art methods in both quantitative and qualitative metrics.

\begin{IEEEbiographynophoto}{Jane Doe}
Biography text here without a photo.
\end{IEEEbiographynophoto}

% \begin{IEEEbiography}[{\includegraphics[width=1in,height=1.25in,clip,keepaspectratio]{fig1.png}}]{IEEE Publications Technology Team}
% In this paragraph you can place your educational, professional background and research and other interests.\end{IEEEbiography}


\begin{thebibliography}{10}
\providecommand{\url}[1]{#1}
\csname url@samestyle\endcsname
\providecommand{\newblock}{\relax}
\providecommand{\bibinfo}[2]{#2}
\providecommand{\BIBentrySTDinterwordspacing}{\spaceskip=0pt\relax}
\providecommand{\BIBentryALTinterwordstretchfactor}{4}
\providecommand{\BIBentryALTinterwordspacing}{\spaceskip=\fontdimen2\font plus
\BIBentryALTinterwordstretchfactor\fontdimen3\font minus
  \fontdimen4\font\relax}
\providecommand{\BIBforeignlanguage}[2]{{%
\expandafter\ifx\csname l@#1\endcsname\relax
\typeout{** WARNING: IEEEtran.bst: No hyphenation pattern has been}%
\typeout{** loaded for the language `#1'. Using the pattern for}%
\typeout{** the default language instead.}%
\else
\language=\csname l@#1\endcsname
\fi
#2}}
\providecommand{\BIBdecl}{\relax}
\BIBdecl

\bibitem{song2022medical}
X.~Song, X.-J. Wu, and H.~Li, ``A medical image fusion method based on
  mdlatlrrv2,'' \emph{arXiv preprint arXiv:2206.15179}, 2022.

\bibitem{zang2021ufa}
Y.~Zang, D.~Zhou, C.~Wang, R.~Nie, and Y.~Guo, ``Ufa-fuse: A novel deep
  supervised and hybrid model for multifocus image fusion,'' \emph{IEEE
  Transactions on Instrumentation and Measurement}, vol.~70, pp. 1--17, 2021.

\bibitem{liu2022attention}
J.~Liu, J.~Shang, R.~Liu, and X.~Fan, ``Attention-guided global-local
  adversarial learning for detail-preserving multi-exposure image fusion,''
  \emph{IEEE Transactions on Circuits and Systems for Video Technology},
  vol.~32, no.~8, pp. 5026--5040, 2022.

\bibitem{ma2019infrared}
J.~Ma, Y.~Ma, and C.~Li, ``Infrared and visible image fusion methods and
  applications: A survey,'' \emph{Information Fusion}, vol.~45, pp. 153--178,
  2019.

\bibitem{zhang2020vifb}
X.~Zhang, P.~Ye, and G.~Xiao, ``Vifb: A visible and infrared image fusion
  benchmark,'' in \emph{Proceedings of the IEEE/CVF Conference on Computer
  Vision and Pattern Recognition Workshops}, 2020, pp. 104--105.

\bibitem{cao2019pedestrian}
Y.~Cao, D.~Guan, W.~Huang, J.~Yang, Y.~Cao, and Y.~Qiao, ``Pedestrian detection
  with unsupervised multispectral feature learning using deep neural
  networks,'' \emph{information fusion}, vol.~46, pp. 206--217, 2019.

\bibitem{liu2022target}
J.~Liu, X.~Fan, Z.~Huang, G.~Wu, R.~Liu, W.~Zhong, and Z.~Luo, ``Target-aware
  dual adversarial learning and a multi-scenario multi-modality benchmark to
  fuse infrared and visible for object detection,'' in \emph{Proceedings of the
  IEEE/CVF Conference on Computer Vision and Pattern Recognition}, 2022, pp.
  5802--5811.

\bibitem{zhang2020object}
X.~Zhang, P.~Ye, H.~Leung, K.~Gong, and G.~Xiao, ``Object fusion tracking based
  on visible and infrared images: A comprehensive review,'' \emph{Information
  Fusion}, vol.~63, pp. 166--187, 2020.

\bibitem{zeng2023random}
X.~Zeng, J.~Long, S.~Tian, and G.~Xiao, ``Random area pixel variation and
  random area transform for visible-infrared cross-modal pedestrian
  re-identification,'' \emph{Expert Systems with Applications}, vol. 215, p.
  119307, 2023.

\bibitem{hou2021generative}
J.~Hou, D.~Zhang, W.~Wu, J.~Ma, and H.~Zhou, ``A generative adversarial network
  for infrared and visible image fusion based on semantic segmentation,''
  \emph{Entropy}, vol.~23, no.~3, p. 376, 2021.

\bibitem{gan2015infrared}
W.~Gan, X.~Wu, W.~Wu, X.~Yang, C.~Ren, X.~He, and K.~Liu, ``Infrared and
  visible image fusion with the use of multi-scale edge-preserving
  decomposition and guided image filter,'' \emph{Infrared Physics \&
  Technology}, vol.~72, pp. 37--51, 2015.

\bibitem{zhu2018novel}
Z.~Zhu, H.~Yin, Y.~Chai, Y.~Li, and G.~Qi, ``A novel multi-modality image
  fusion method based on image decomposition and sparse representation,''
  \emph{Information Sciences}, vol. 432, pp. 516--529, 2018.

\bibitem{li2020mdlatlrr}
H.~Li, X.-J. Wu, and J.~Kittler, ``Mdlatlrr: A novel decomposition method for
  infrared and visible image fusion,'' \emph{IEEE Transactions on Image
  Processing}, vol.~29, pp. 4733--4746, 2020.

\bibitem{bavirisetti2016two}
D.~P. Bavirisetti and R.~Dhuli, ``Two-scale image fusion of visible and
  infrared images using saliency detection,'' \emph{Infrared Physics \&
  Technology}, vol.~76, pp. 52--64, 2016.

\bibitem{zhang2015fusion}
B.~Zhang, X.~Lu, H.~Pei, and Y.~Zhao, ``A fusion algorithm for infrared and
  visible images based on saliency analysis and non-subsampled shearlet
  transform,'' \emph{Infrared Physics \& Technology}, vol.~73, pp. 286--297,
  2015.

\bibitem{liu2016image}
Y.~Liu, X.~Chen, R.~K. Ward, and Z.~J. Wang, ``Image fusion with convolutional
  sparse representation,'' \emph{IEEE signal processing letters}, vol.~23,
  no.~12, pp. 1882--1886, 2016.

\bibitem{yang2020infrared}
Y.~Yang, Y.~Zhang, S.~Huang, Y.~Zuo, and J.~Sun, ``Infrared and visible image
  fusion using visual saliency sparse representation and detail injection
  model,'' \emph{IEEE Transactions on Instrumentation and Measurement},
  vol.~70, pp. 1--15, 2020.

\bibitem{ma2016infrared}
J.~Ma, C.~Chen, C.~Li, and J.~Huang, ``Infrared and visible image fusion via
  gradient transfer and total variation minimization,'' \emph{Information
  Fusion}, vol.~31, pp. 100--109, 2016.

\bibitem{madheswari2017swarm}
K.~Madheswari and N.~Venkateswaran, ``Swarm intelligence based optimisation in
  thermal image fusion using dual tree discrete wavelet transform,''
  \emph{Quantitative Infrared Thermography Journal}, vol.~14, no.~1, pp.
  24--43, 2017.

\bibitem{ma2017infrared}
J.~Ma, Z.~Zhou, B.~Wang, and H.~Zong, ``Infrared and visible image fusion based
  on visual saliency map and weighted least square optimization,''
  \emph{Infrared Physics \& Technology}, vol.~82, pp. 8--17, 2017.

\bibitem{xu2022asymmetric}
C.~Xu, Q.~Li, Q.~Zhou, X.~Jiang, D.~Yu, and Y.~Zhou, ``Asymmetric cross-modal
  activation network for rgb-t salient object detection,''
  \emph{Knowledge-Based Systems}, vol. 258, p. 110047, 2022.

\bibitem{wani2022lowlight}
P.~Wani, K.~Usmani, G.~Krishnan, T.~O’Connor, and B.~Javidi, ``Lowlight
  object recognition by deep learning with passive three-dimensional integral
  imaging in visible and long wave infrared wavelengths,'' \emph{Optics
  Express}, vol.~30, no.~2, pp. 1205--1218, 2022.

\bibitem{wang2022freesolo}
X.~Wang, Z.~Yu, S.~De~Mello, J.~Kautz, A.~Anandkumar, C.~Shen, and J.~M.
  Alvarez, ``Freesolo: Learning to segment objects without annotations,'' in
  \emph{Proceedings of the IEEE/CVF Conference on Computer Vision and Pattern
  Recognition}, 2022, pp. 14\,176--14\,186.

\bibitem{shen2022learning}
X.~Shen, A.~A. Efros, A.~Joulin, and M.~Aubry, ``Learning co-segmentation by
  segment swapping for retrieval and discovery,'' in \emph{Proceedings of the
  IEEE/CVF Conference on Computer Vision and Pattern Recognition}, 2022, pp.
  5082--5092.

\bibitem{yuan2020scale}
Y.~Yuan, J.~Chu, L.~Leng, J.~Miao, and B.-G. Kim, ``A scale-adaptive
  object-tracking algorithm with occlusion detection,'' \emph{EURASIP Journal
  on Image and Video Processing}, vol. 2020, pp. 1--15, 2020.

\bibitem{wang2021adaptive}
Y.~Wang, X.~Wei, X.~Tang, H.~Shen, and H.~Zhang, ``Adaptive fusion cnn features
  for rgbt object tracking,'' \emph{IEEE Transactions on Intelligent
  Transportation Systems}, vol.~23, no.~7, pp. 7831--7840, 2021.

\bibitem{liu2021learning}
J.~Liu, X.~Fan, J.~Jiang, R.~Liu, and Z.~Luo, ``Learning a deep multi-scale
  feature ensemble and an edge-attention guidance for image fusion,''
  \emph{IEEE Transactions on Circuits and Systems for Video Technology},
  vol.~32, no.~1, pp. 105--119, 2021.

\bibitem{jian2020sedrfuse}
L.~Jian, X.~Yang, Z.~Liu, G.~Jeon, M.~Gao, and D.~Chisholm, ``Sedrfuse: A
  symmetric encoder--decoder with residual block network for infrared and
  visible image fusion,'' \emph{IEEE Transactions on Instrumentation and
  Measurement}, vol.~70, pp. 1--15, 2020.

\bibitem{xu2021drf}
H.~Xu, X.~Wang, and J.~Ma, ``Drf: Disentangled representation for visible and
  infrared image fusion,'' \emph{IEEE Transactions on Instrumentation and
  Measurement}, vol.~70, pp. 1--13, 2021.

\bibitem{tang2022image}
L.~Tang, J.~Yuan, and J.~Ma, ``Image fusion in the loop of high-level vision
  tasks: A semantic-aware real-time infrared and visible image fusion
  network,'' \emph{Information Fusion}, vol.~82, pp. 28--42, 2022.

\bibitem{long2021rxdnfuse}
Y.~Long, H.~Jia, Y.~Zhong, Y.~Jiang, and Y.~Jia, ``Rxdnfuse: A aggregated
  residual dense network for infrared and visible image fusion,''
  \emph{Information Fusion}, vol.~69, pp. 128--141, 2021.

\bibitem{li2021different}
H.~Li, Y.~Cen, Y.~Liu, X.~Chen, and Z.~Yu, ``Different input resolutions and
  arbitrary output resolution: A meta learning-based deep framework for
  infrared and visible image fusion,'' \emph{IEEE Transactions on Image
  Processing}, vol.~30, pp. 4070--4083, 2021.

\bibitem{ma2019fusiongan}
J.~Ma, W.~Yu, P.~Liang, C.~Li, and J.~Jiang, ``Fusiongan: A generative
  adversarial network for infrared and visible image fusion,''
  \emph{Information fusion}, vol.~48, pp. 11--26, 2019.

\bibitem{zhou2021semantic}
H.~Zhou, W.~Wu, Y.~Zhang, J.~Ma, and H.~Ling, ``Semantic-supervised infrared
  and visible image fusion via a dual-discriminator generative adversarial
  network,'' \emph{IEEE Transactions on Multimedia}, 2021.

\bibitem{xu2019learning}
H.~Xu, P.~Liang, W.~Yu, J.~Jiang, and J.~Ma, ``Learning a generative model for
  fusing infrared and visible images via conditional generative adversarial
  network with dual discriminators.'' in \emph{IJCAI}, 2019, pp. 3954--3960.

\bibitem{xu2021classification}
H.~Xu, H.~Zhang, and J.~Ma, ``Classification saliency-based rule for visible
  and infrared image fusion,'' \emph{IEEE Transactions on Computational
  Imaging}, vol.~7, pp. 824--836, 2021.

\bibitem{liu2020bilevel}
R.~Liu, J.~Liu, Z.~Jiang, X.~Fan, and Z.~Luo, ``A bilevel integrated model with
  data-driven layer ensemble for multi-modality image fusion,'' \emph{IEEE
  Transactions on Image Processing}, vol.~30, pp. 1261--1274, 2020.

\bibitem{ma2020ddcgan}
J.~Ma, H.~Xu, J.~Jiang, X.~Mei, and X.-P. Zhang, ``Ddcgan: A dual-discriminator
  conditional generative adversarial network for multi-resolution image
  fusion,'' \emph{IEEE Transactions on Image Processing}, vol.~29, pp.
  4980--4995, 2020.

\bibitem{gao2022dcdr}
Y.~Gao, S.~Ma, and J.~Liu, ``Dcdr-gan: A densely connected disentangled
  representation generative adversarial network for infrared and visible image
  fusion,'' \emph{IEEE Transactions on Circuits and Systems for Video
  Technology}, vol.~33, no.~2, pp. 549--561, 2022.

\bibitem{yue2023dif}
J.~Yue, L.~Fang, S.~Xia, Y.~Deng, and J.~Ma, ``Dif-fusion: Towards high color
  fidelity in infrared and visible image fusion with diffusion models,''
  \emph{arXiv preprint arXiv:2301.08072}, 2023.

\bibitem{woo2018cbam}
S.~Woo, J.~Park, J.-Y. Lee, and I.~S. Kweon, ``Cbam: Convolutional block
  attention module,'' in \emph{Proceedings of the European conference on
  computer vision (ECCV)}, 2018, pp. 3--19.

\bibitem{wang2023fusiongram}
J.~Wang, X.~Xi, D.~Li, and F.~Li, ``Fusiongram: An infrared and visible image
  fusion framework based on gradient residual and attention mechanism,''
  \emph{IEEE Transactions on Instrumentation and Measurement}, vol.~72, pp.
  1--12, 2023.

\bibitem{zheng2022multi}
X.~Zheng, Q.~Yang, P.~Si, and Q.~Wu, ``A multi-stage visible and infrared image
  fusion network based on attention mechanism,'' \emph{Sensors}, vol.~22,
  no.~10, p. 3651, 2022.

\bibitem{meng2022multilayer}
Q.~Meng, M.~Zhao, L.~Zhang, W.~Shi, C.~Su, and L.~Bruzzone, ``Multilayer
  feature fusion network with spatial attention and gated mechanism for remote
  sensing scene classification,'' \emph{IEEE Geoscience and Remote Sensing
  Letters}, vol.~19, pp. 1--5, 2022.

\bibitem{wang2020self}
Y.~Wang, J.~Zhang, M.~Kan, S.~Shan, and X.~Chen, ``Self-supervised equivariant
  attention mechanism for weakly supervised semantic segmentation,'' in
  \emph{Proceedings of the IEEE/CVF conference on computer vision and pattern
  recognition}, 2020, pp. 12\,275--12\,284.

\bibitem{zhao2021semantic}
Q.~Zhao, J.~Liu, Y.~Li, and H.~Zhang, ``Semantic segmentation with attention
  mechanism for remote sensing images,'' \emph{IEEE Transactions on Geoscience
  and Remote Sensing}, vol.~60, pp. 1--13, 2021.

\bibitem{xu2020u2fusion}
H.~Xu, J.~Ma, J.~Jiang, X.~Guo, and H.~Ling, ``U2fusion: A unified unsupervised
  image fusion network,'' \emph{IEEE Transactions on Pattern Analysis and
  Machine Intelligence}, 2020.

\bibitem{li2011performance}
S.~Li, B.~Yang, and J.~Hu, ``Performance comparison of different
  multi-resolution transforms for image fusion,'' \emph{Information Fusion},
  vol.~12, no.~2, pp. 74--84, 2011.

\bibitem{lewis2007pixel}
J.~J. Lewis, R.~J. O’Callaghan, S.~G. Nikolov, D.~R. Bull, and
  N.~Canagarajah, ``Pixel-and region-based image fusion with complex
  wavelets,'' \emph{Information fusion}, vol.~8, no.~2, pp. 119--130, 2007.

\bibitem{zhang2017infrared}
X.~Zhang, Y.~Ma, F.~Fan, Y.~Zhang, and J.~Huang, ``Infrared and visible image
  fusion via saliency analysis and local edge-preserving multi-scale
  decomposition,'' \emph{JOSA A}, vol.~34, no.~8, pp. 1400--1410, 2017.

\bibitem{liu2015general}
Y.~Liu, S.~Liu, and Z.~Wang, ``A general framework for image fusion based on
  multi-scale transform and sparse representation,'' \emph{Information fusion},
  vol.~24, pp. 147--164, 2015.

\bibitem{vanmali2017visible}
A.~V. Vanmali and V.~M. Gadre, ``Visible and nir image fusion using
  weight-map-guided laplacian--gaussian pyramid for improving scene
  visibility,'' \emph{S{\=a}dhan{\=a}}, vol.~42, pp. 1063--1082, 2017.

\bibitem{ren2022fusion}
Z.~Ren, G.~Ren, and D.~Wu, ``Fusion of infrared and visible images based on
  discrete cosine wavelet transform and high pass filter,'' \emph{Soft
  Computing}, pp. 1--12, 2022.

\bibitem{li2013image}
S.~Li, X.~Kang, and J.~Hu, ``Image fusion with guided filtering,'' \emph{IEEE
  Transactions on Image processing}, vol.~22, no.~7, pp. 2864--2875, 2013.

\bibitem{liu2017infrared}
C.~Liu, Y.~Qi, and W.~Ding, ``Infrared and visible image fusion method based on
  saliency detection in sparse domain,'' \emph{Infrared Physics \& Technology},
  vol.~83, pp. 94--102, 2017.

\bibitem{li2023infrared}
Y.~Li, G.~Liu, D.~P. Bavirisetti, X.~Gu, and X.~Zhou, ``Infrared-visible image
  fusion method based on sparse and prior joint saliency detection and
  latlrr-fpde,'' \emph{Digital Signal Processing}, p. 103910, 2023.

\bibitem{yang2023latlrr}
Y.~Yang, C.~Gao, Z.~Ming, J.~Guo, E.~Leopold, J.~Cheng, J.~Zuo, and M.~Zhu,
  ``Latlrr-cnn: an infrared and visible image fusion method combining latent
  low-rank representation and cnn,'' \emph{Multimedia Tools and Applications},
  pp. 1--21, 2023.

\bibitem{zhang2018sparse}
Q.~Zhang, Y.~Liu, R.~S. Blum, J.~Han, and D.~Tao, ``Sparse representation based
  multi-sensor image fusion for multi-focus and multi-modality images: A
  review,'' \emph{Information Fusion}, vol.~40, pp. 57--75, 2018.

\bibitem{ding2018infrared}
W.~Ding, D.~Bi, L.~He, and Z.~Fan, ``Infrared and visible image fusion method
  based on sparse features,'' \emph{Infrared Physics \& Technology}, vol.~92,
  pp. 372--380, 2018.

\bibitem{li2018infrared}
H.~Li, X.-J. Wu, and J.~Kittler, ``Infrared and visible image fusion using a
  deep learning framework,'' in \emph{2018 24th international conference on
  pattern recognition (ICPR)}.\hskip 1em plus 0.5em minus 0.4em\relax IEEE,
  2018, pp. 2705--2710.

\bibitem{zhang2022infrared}
L.~Zhang, H.~Li, R.~Zhu, and P.~Du, ``An infrared and visible image fusion
  algorithm based on resnet-152,'' \emph{Multimedia Tools and Applications},
  vol.~81, no.~7, pp. 9277--9287, 2022.

\bibitem{li2020nestfuse}
H.~Li, X.-J. Wu, and T.~Durrani, ``Nestfuse: An infrared and visible image
  fusion architecture based on nest connection and spatial/channel attention
  models,'' \emph{IEEE Transactions on Instrumentation and Measurement},
  vol.~69, no.~12, pp. 9645--9656, 2020.

\bibitem{li2018densefuse}
H.~Li and X.-J. Wu, ``Densefuse: A fusion approach to infrared and visible
  images,'' \emph{IEEE Transactions on Image Processing}, vol.~28, no.~5, pp.
  2614--2623, 2018.

\bibitem{lin2014microsoft}
T.-Y. Lin, M.~Maire, S.~Belongie, J.~Hays, P.~Perona, D.~Ramanan,
  P.~Doll{\'a}r, and C.~L. Zitnick, ``Microsoft coco: Common objects in
  context,'' in \emph{Computer Vision--ECCV 2014: 13th European Conference,
  Zurich, Switzerland, September 6-12, 2014, Proceedings, Part V 13}.\hskip 1em
  plus 0.5em minus 0.4em\relax Springer, 2014, pp. 740--755.

\bibitem{xu2022cufd}
H.~Xu, M.~Gong, X.~Tian, J.~Huang, and J.~Ma, ``Cufd: An encoder--decoder
  network for visible and infrared image fusion based on common and unique
  feature decomposition,'' \emph{Computer Vision and Image Understanding}, vol.
  218, p. 103407, 2022.

\bibitem{huang2017densely}
G.~Huang, Z.~Liu, L.~Van Der~Maaten, and K.~Q. Weinberger, ``Densely connected
  convolutional networks,'' in \emph{Proceedings of the IEEE conference on
  computer vision and pattern recognition}, 2017, pp. 4700--4708.

\bibitem{xu2022multi}
D.~Xu, N.~Zhang, Y.~Zhang, Z.~Li, Z.~Zhao, and Y.~Wang, ``Multi-scale
  unsupervised network for infrared and visible image fusion based on joint
  attention mechanism,'' \emph{Infrared Physics \& Technology}, vol. 125, p.
  104242, 2022.

\bibitem{wang2004image}
Z.~Wang, A.~C. Bovik, H.~R. Sheikh, and E.~P. Simoncelli, ``Image quality
  assessment: from error visibility to structural similarity,'' \emph{IEEE
  transactions on image processing}, vol.~13, no.~4, pp. 600--612, 2004.

\bibitem{zhou2022unified}
H.~Zhou, J.~Hou, Y.~Zhang, J.~Ma, and H.~Ling, ``Unified gradient-and
  intensity-discriminator generative adversarial network for image fusion,''
  \emph{Information Fusion}, vol.~88, pp. 184--201, 2022.

\bibitem{xu2020deep}
H.~Xu, F.~Fan, H.~Zhang, Z.~Le, and J.~Huang, ``A deep model for multi-focus
  image fusion based on gradients and connected regions,'' \emph{IEEE Access},
  vol.~8, pp. 26\,316--26\,327, 2020.

\bibitem{ram2017deepfuse}
K.~Ram~Prabhakar, V.~Sai~Srikar, and R.~Venkatesh~Babu, ``Deepfuse: A deep
  unsupervised approach for exposure fusion with extreme exposure image
  pairs,'' in \emph{Proceedings of the IEEE international conference on
  computer vision}, 2017, pp. 4714--4722.

\bibitem{li2019infrared}
H.~Li, X.-j. Wu, and T.~S. Durrani, ``Infrared and visible image fusion with
  resnet and zero-phase component analysis,'' \emph{Infrared Physics \&
  Technology}, vol. 102, p. 103039, 2019.

\bibitem{zhao2020didfuse}
Z.~Zhao, S.~Xu, C.~Zhang, J.~Liu, P.~Li, and J.~Zhang, ``Didfuse: Deep image
  decomposition for infrared and visible image fusion,'' \emph{arXiv preprint
  arXiv:2003.09210}, 2020.

\bibitem{xu2020fusiondn}
H.~Xu, J.~Ma, Z.~Le, J.~Jiang, and X.~Guo, ``Fusiondn: A unified densely
  connected network for image fusion,'' in \emph{Proceedings of the AAAI
  Conference on Artificial Intelligence}, vol.~34, no.~07, 2020, pp.
  12\,484--12\,491.

\bibitem{toet2012progress}
A.~Toet and M.~A. Hogervorst, ``Progress in color night vision,'' \emph{Optical
  Engineering}, vol.~51, no.~1, p. 010901, 2012.

\bibitem{roberts2008assessment}
J.~W. Roberts, J.~A. Van~Aardt, and F.~B. Ahmed, ``Assessment of image fusion
  procedures using entropy, image quality, and multispectral classification,''
  \emph{Journal of Applied Remote Sensing}, vol.~2, no.~1, p. 023522, 2008.

\bibitem{qu2002information}
G.~Qu, D.~Zhang, and P.~Yan, ``Information measure for performance of image
  fusion,'' \emph{Electronics letters}, vol.~38, no.~7, p.~1, 2002.

\bibitem{wang2002universal}
Z.~Wang and A.~C. Bovik, ``A universal image quality index,'' \emph{IEEE signal
  processing letters}, vol.~9, no.~3, pp. 81--84, 2002.

\bibitem{wu2005remote}
J.~Wu, H.~Huang, Y.~Qiu, H.~Wu, J.~Tian, and J.~Liu, ``Remote sensing image
  fusion based on average gradient of wavelet transform,'' in \emph{IEEE
  International Conference Mechatronics and Automation, 2005}, vol.~4.\hskip
  1em plus 0.5em minus 0.4em\relax IEEE, 2005, pp. 1817--1821.

\bibitem{rao1997fibre}
Y.-J. Rao, ``In-fibre bragg grating sensors,'' \emph{Measurement science and
  technology}, vol.~8, no.~4, p. 355, 1997.

\bibitem{eskicioglu1995image}
A.~M. Eskicioglu and P.~S. Fisher, ``Image quality measures and their
  performance,'' \emph{IEEE Transactions on communications}, vol.~43, no.~12,
  pp. 2959--2965, 1995.

\bibitem{xydeas2000objective}
C.~S. Xydeas, V.~Petrovic \emph{et~al.}, ``Objective image fusion performance
  measure,'' \emph{Electronics letters}, vol.~36, no.~4, pp. 308--309, 2000.

\bibitem{han2013new}
Y.~Han, Y.~Cai, Y.~Cao, and X.~Xu, ``A new image fusion performance metric
  based on visual information fidelity,'' \emph{Information fusion}, vol.~14,
  no.~2, pp. 127--135, 2013.

\bibitem{ma2021stdfusionnet}
J.~Ma, L.~Tang, M.~Xu, H.~Zhang, and G.~Xiao, ``Stdfusionnet: An infrared and
  visible image fusion network based on salient target detection,'' \emph{IEEE
  Transactions on Instrumentation and Measurement}, vol.~70, pp. 1--13, 2021.

\bibitem{wan2021lightweight}
H.~Wan, J.~Chen, Z.~Huang, Y.~Feng, Z.~Zhou, X.~Liu, B.~Yao, and T.~Xu,
  ``Lightweight channel attention and multiscale feature fusion discrimination
  for remote sensing scene classification,'' \emph{IEEE Access}, vol.~9, pp.
  94\,586--94\,600, 2021.

\bibitem{xiao2023mfmanet}
H.~Xiao, Q.~Liu, and L.~Li, ``Mfmanet: Multi-feature multi-attention network
  for efficient subtype classification on non-small cell lung cancer ct
  images,'' \emph{Biomedical Signal Processing and Control}, vol.~84, p.
  104768, 2023.

\end{thebibliography}
\end{document}